\documentclass{article}

\usepackage{microtype}
\usepackage{graphicx}
\usepackage{subfigure}
\usepackage{booktabs} 
\usepackage{amsmath,amssymb,mathrsfs}
\usepackage{etoolbox, makecell, fancyvrb}
\usepackage{hyperref}

\usepackage[accepted]{arxiv}
\usepackage{enumitem}
\setlist{nolistsep}

\newcommand{\cd}[1]{\texttt{\small #1}}
\newcommand{\pprank}{\texttt{pp\_rank}}
\newcommand{\tprank}{\texttt{tp\_rank}}
\newcommand{\dprank}{\texttt{dp\_rank} }

\newcommand{\rank}{\texttt{rank} }

\newcommand{\ppgroup}{$\mathsf{PP\_GROUP}$}
\newcommand{\tpgroup}{$\mathsf{TP\_GROUP}$}
\newcommand{\dpgroup}{$\mathsf{DP\_GROUP}$}

\newcommand{\rdpgroup}{$\mathsf{RDP\_GROUP}$}
\makeatletter
\patchcmd{\@verbatim}
  {\verbatim@font}
  {\verbatim@font\footnotesize}
  {}{}
\makeatother

\newcommand{\lp}{\left(}
\newcommand{\rp}{\right)}
\newcommand{\lb}{\left[}
\newcommand{\rb}{\right]}
\newcommand{\lbp}{\left\{}
\newcommand{\rbp}{\right\}}

\mlsystitlerunning{Amazon SageMaker Model Parallelism: A General and Flexible Framework for Large Model Training}

\begin{document}

\twocolumn[
\mlsystitle{Amazon SageMaker Model Parallelism: A General and Flexible Framework for Large Model Training}



\mlsyssetsymbol{equal}{*}

\begin{mlsysauthorlist}
\mlsysauthor{Can Karakus}{amzn}
\mlsysauthor{Rahul Huilgol}{amzn}
\mlsysauthor{Fei Wu}{amzn}
\mlsysauthor{Anirudh Subramanian}{amzn}
\mlsysauthor{Cade Daniel}{amzn}
\mlsysauthor{Derya Cavdar}{amzn}
\mlsysauthor{Teng Xu}{amzn}
\mlsysauthor{Haohan Chen}{amzn}
\mlsysauthor{Arash Rahnama}{amzn}
\mlsysauthor{Luis Quintela}{amzn}
\end{mlsysauthorlist}

\mlsysaffiliation{amzn}{Amazon.com, Santa Clara, CA}

\mlsyscorrespondingauthor{Can Karakus}{cakarak@amazon.com}

\mlsyskeywords{Machine Learning, MLSys}

\vskip 0.3in

\begin{abstract}
With deep learning models rapidly growing in size, systems-level solutions for large-model training are required. We present Amazon SageMaker model parallelism, a software library that integrates with PyTorch, and enables easy training of large models using model parallelism and other memory-saving features. In contrast to existing solutions, the implementation of the SageMaker library is much more generic and flexible, in that it can automatically partition and run pipeline parallelism over arbitrary model architectures with minimal code change, and also offers a general and extensible framework for tensor parallelism, which supports a wider range of use cases, and is modular enough to be easily applied to new training scripts. The library also preserves the native PyTorch user experience to a much larger degree, supporting module re-use and dynamic graphs, while giving the user full control over the details of the training step. We evaluate performance over GPT-3, RoBERTa, BERT, and neural collaborative filtering, and demonstrate competitive performance over existing solutions.
\end{abstract}
]



\printAffiliationsAndNotice{\mlsysEqualContribution} 

\section{Introduction}\label{sec:introduction}
Recent years have seen an exponential increase in the size of the state-of-the-art deep learning models, measured in the number of trainable parameters, 
driven by the observation that larger models achieve better generalization performance, as well as demonstrating examples of zero-shot and few-shot generalization behaviors \cite{BrownMann_20}.
This trend has spurred interest in systems-level solutions for large model training, since the model sizes far outgrew the available memory capacity in state-of-the-art hardware accelerators. 
Such solutions consist of partitioning the model parameters and other memory-consuming training state (gradients, optimizer states, activations) across devices (model parallelism), as well as other memory-saving techniques.



Although the existing model parallelism solutions have been successful in some applications, there remains a need for a generic framework that can flexibly handle the full diversity of possible use cases. This is because the existing solutions for the two types of model parallelism, namely pipeline parallelism and tensor parallelism, are typically limited either in the supported use cases, model architectures, or framework APIs/features; or require a prohibitively large effort to integrate with a new training script. Some unsupported use cases might include, but are not limited to, (1) architectures that do not consist of a single transformer encoder (multiple transformer blocks such as T5 \cite{RaffelShazeer_19}, RAG \cite{LewisPerez_20}, REALM \cite{GuuLee_20}, etc. or large non-transformer architectures \cite{RealAggarwal_19}), (2) architectures that do not consist of a consecutive sequence of identical layers, (3) a single large component in an otherwise small model (a huge output layer due to too many classes, or a huge embedding layer, especially in recommendation models with billions of items), (4) architectures that make extensive use of module/parameter re-use, (5) scripts with conditional execution flows, (6) non-conventional execution patterns such as mixture-of-experts layers.

In this paper, we present a general, flexible, and extensible framework for large model training, which includes both pipeline and tensor parallelism, as well as other popular memory-saving features, and covers all of the above-mentioned use cases \emph{in addition to} the commonly-studied single-transformer architectures. The library is designed to take minimal effort to integrate with a brand new script, regardless of model architecture and the API used. The library automatically partitions and distributes \emph{arbitrary} model architectures across multiple devices (possibly across nodes) without explicit user input on how to partition (see \S\ref{sec:ux} for on overview of the API), and internally manages training runtime, including pipelining and cross-device communication through its dedicated communication backend. In addition to being generic, our experiments also show that the library has competitive performance with respect to DeepSpeed.

Our main contributions are as follows:
\begin{itemize}
\item Design and implementation of a pipeline parallelism engine, including a load-balancing auto-partitioning algorithm and pipelining runtime for arbitrary model architectures based on \emph{module-server} design,
\item A general and extensible tensor parallelism framework that is applicable to wider range of scenarios than existing solutions, including uniformly large models, models with a limited number of large components, and mixture-of-experts models \cite{ShazeerMirhoseini_17}, along with built-in distributed module implementations for commonly-used building blocks in deep learning,
\item A dedicated device-to-device (D2D) communication backend design that can handle dynamically-generated communication requests,
\item A flexible user API design that does not abstract away the details of training step, and maintains most of the underlying framework features and characteristics such as module re-use and dynamic graphs,
\item A set of experiments on AWS infrastructure to train GPT-3 \cite{BrownMann_20}, BERT \cite{DevlinChang_18}, RoBERTa \cite{LiuOtt_19}, and Neural Collaborative Filtering (NCF) \cite{HeLiao_17} models, which demonstrates the performance of the library.
\end{itemize}
The rest of the paper is organized as follows. We review relevant literature in \S\ref{sec:relatedwork}, present the design, overview, and the API of the library in \S\ref{sec:design}, describe pipeline parallelism architecture in \S\ref{sec:pipeline} and tensor parallelism architecture in \S\ref{sec:tensor}, explain the design of the communication backend in \S\ref{sec:backend}, and present the empirical results in \S\ref{sec:experiments}. 

The paper assumes a certain familiarity with model parallelism concepts such as pipeline parallelism, tensor parallelism, and microbatching. For a primer on such topics, refer to Appendix~\ref{sec:ap_modelparallel}.

\section{Related Work}\label{sec:relatedwork}
In recent years, there has been increasing interest in model parallelism and other large model training solutions. Among the first was GPipe \cite{HuangCheng_19} and PipeDream \cite{NarayananHarlap_19}, the latter of which improves pipeline efficiency at the expense of increased memory use due to storing multiple weight copies. The works in \cite{ChenYang_18} and \cite{NarayananPhanishayee_21} aimed to address this issue through weight prediction, and a novel weight update scheme, respectively. TeraPipe \cite{LiZhuang_21} introduced another type of pipelining specific to single-transformer architectures, where pipelining occurs across tokens instead of microbatches.


Another type of model parallelism is \emph{tensor parallelism}, where individual operators or layers are partitioned. Mesh-TensorFlow (MTF) \cite{ShazeerCheng_18} created a tensor parallelism framework on top of TensorFlow. Megatron-LM \cite{ShoeybiPatwary_19} created a tensor-parallel implementation of GPT-2 and T5 based on PyTorch, and added pipeline parallelism in later work \cite{NarayananShoeybi_21}. More recently, GSPMD \cite{XuLee_21} implemented tensor parallelism as part of XLA compiler. Partitioning over sequence dimension for transformers was proposed in \cite{LiXue_21}.

A separate line of work in scaling up deep learning models has been based on the mixture-of-experts (MoE) layer \cite{ShazeerMirhoseini_17}, which scaled models by having extremely wide layers consisting of parallel ``experts", and forwarding incoming activations to only $k$ experts using a differentiable gating mechanism. GShard \cite{LepikhinLee_20} implemented this idea in XLA compiler to train a 600 billion parameter model. Switch Transformers \cite{FedusZoph_21} built upon this idea, forward each input to only one expert, and further scaling the achievable model scale.


The DeepSpeed project produced a line of work that combined a number of large-model training techniques \cite{RasleyRajbhandari_20, RajbhandariRasley_20, RenRajbhandari_21, RajbhandariRuwase_21}. This was centered around the ZeRO optimizer, which shards the optimizer states, gradients, and parameters across data-parallel devices. Later, CPU offloading techniques were also added to enable multi-trillion-parameter model training \cite{RenRajbhandari_21}.

In contrast to these works, this paper presents a large-model training solution that is more generic, architecture-agnostic, flexible, and easy-to-use. For instance, the pipeline and tensor parallelism implementations of Megatron-LM are deeply integrated with GPT-3 model definition code, and not readily applicable to a new training script, a novel architecture, or a new training technique that requires direct access to loss or gradient tensors. Similarly, the popular DeepSpeed library requires significant changes to the training script to implement pipeline parallelism, requires \cd{nn.Sequential} API to be used, and hides the details of the training step under a high-level API, taking control away from user. Further, although the sharding techniques of ZeRO optimizer are often very effective in achieving large scale training (and are partially also implemented in SageMaker model parallelism), under some scenarios, such as the use of novel unsupported optimizers, or very large embeddings in recommendation models, sharding techniques might become infeasible or less performant than model parallelism. Our library addresses all such limitations, as will be explained in next section.



\section{Design Principles and Overview}
\label{sec:design}
\subsection{Design principles}
Amazon SageMaker model parallelism library is designed to be generic, flexible, and easy-to-use, which means that the library can be integrated into an existing training script with a small number of additional lines of code, under a wide range of scenarios. Some specific ways in which the library provides this flexibility are as follows:
\begin{enumerate}
\item {\bf Do not abstract away the training step: } Deep learning literature contains a wide array of useful training techniques, and the training of state-of-the-art models often benefits from such techniques in terms of accuracy and convergence. These techniques may include gradient clipping, mixed-precision training with loss scaling, or use of various novel optimizers that post-process gradients in differing ways. A critical feature of a generic model parallelism framework must be to allow the user the flexibility to use such techniques as desired, by not abstracting away the details of a training step under a high-level API, and giving explicit access to loss and gradient tensors. This is in general non-trivial to achieve, since pipeline parallelism alters the forward and backward pass in a fundamental manner that does not immediately align with the execution paradigm of the ML framework (currently no existing pipeline parallelism implementation satisfies this to our knowledge). SageMaker model parallelism achieves this by encapsulating the existing training step with a decorator \cd{@smp.step}. The implementation of the training step is still fully exposed (and editable) by the user, while the decorator uses the wrapped function as a building block in the pipelined execution (See \S\ref{sec:ux} and \ref{sec:pipeline} for more details).
\item {\bf Preserve framework features and characteristics: } Modern machine learning frameworks offer a wide range of features that give the user tremendous flexibility in defining models. In contrast, model parallelism solutions built on top of these frameworks are often highly restrictive in terms of which features or APIs are supported. This is because such solutions necessarily have a very large surface area that interacts with the framework, hence supporting the entire set of underlying framework features is a challenge. Despite this, SageMaker model parallelism is designed to maintain most of the core framework features and APIs, so that most existing training scripts can be supported without significant refactors. Some examples of this are as follows:
\begin{itemize}
\item Pipeline parallelism on SageMaker is not limited to particular model definition APIs such as \cd{nn.Sequential}, unlike the existing pipeline parallelism implementations. This allows flexibility to support diverse model architectures that is not limited to a single contiguous block of isomorphic layers, as will be explained in the third point.
\item Module/parameter re-use is automatically handled, without requiring specific input or code change from the user. This is typically a challenge since different versions of a parameter residing on different devices need to be kept synchronized. The library achieves this by placing the modules that share a parameter at the same device automatically, removing the need for synchronization.
\item Conditional execution (\cd{if}/\cd{else} blocks) is supported, thanks to the module-server architecture, which does not have \emph{a priori} assumptions about the architecture or control flow.
\item Training steps with the library are semantically and mathematically equivalent to the original training step, hence it does not introduce model accuracy penalties.
\end{itemize}

\item {\bf Do not limit to specific architectures: } Existing model parallelism solutions typically (almost) exclusively focus on a single model architecture, such as GPT-3. For instance, as of today, Megatron-LM repository is tightly integrated with the GPT-3 model definition and training code, and is not immediately applicable to a generic model architecture. In contrast, pipeline parallelism feature of SageMaker has a generic API that can be readily applied to \emph{any} model architecture in addition to popular transformer architectures. Although tensor parallelism intrinsically requires some degree of architecture specialization, SageMaker tensor parallelism also has a more generic API that is easily applicable to new scripts. It is also applicable under a wider range of scenarios than existing implementations, including large transformers (GPT-3), large embedding tables, or classification models with very large number of classes, owing to the fact that tensor parallelism is applied over \emph{data-parallel} groups. In addition, the API is readily extensible to custom distributed module implementations for modules without built-in support.
\end{enumerate}

\subsection{Framework overview}
\subsubsection{High-level workflow}
The library can be used by making a few lines of code change to an existing training script in PyTorch, and launching a new training job through SageMaker Python SDK. The specific code changes will be discussed in more detail in Section~\ref{sec:ux}. 
Various configurations for model parallelism, as well as other memory-saving features (which include activation checkpointing, activation offloading, and optimizer state sharding, see Appendix~\ref{sec:ap_other} for details), can be defined through a Python \cd{dict} fed through the Estimator API of the SageMaker Python SDK, while launching the job.

Once training is launched, the library automatically traces and partitions the model, and manages the training runtime, including the device placement, pipelined execution, and cross-device communication, both within and across instances. The library also manages data parallelism internally, so the use of a separate data parallelism API is not required. 

Detailed user guide for the library is available in \cite{smp_userguide}, and the detailed API documentation is available in \cite{smp_apidoc}, with supplemental documentation in Appendix~\ref{sec:ap_api}. 

\subsubsection{Process and ranking basics}
The library relies on MPI for process management, and maintains a one-to-one mapping between CPU processes and GPU devices, \emph{i.e.}, each process manages exactly one GPU. For instance, if training is launched over 4 \cd{p4d.24xlarge} instances (with 8 GPUs per instance), an MPI job with 32 processes is launched. We follow the MPI terminology and use \cd{rank} to indicate the global index of the process across the cluster.

\begin{figure}[ht]
\centering
\includegraphics[scale=0.4]{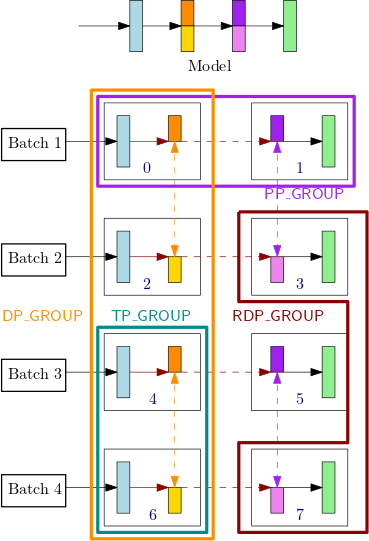}
\caption{Illustration of process groups over 8 devices, with tensor-parallelism degree 2, pipeline-parallelism degree 2, and data-parallelism degree 4. At the top, an example model with 4 layers. On the bottom, the 4-layer model is distributed across 4 devices using both pipeline parallelism and tensor parallelism (tensor parallelism is used for the middle two layers). Note that {\tpgroup} is a \emph{subset} of \dpgroup, since tensor parallelism is performed across data-parallel ranks. Reduced-data-parallel group (\rdpgroup) consists of devices that share identical model replicas.}
\label{fig:ranking}
\end{figure}

The library features three different parallelization strategies; namely, pipeline, tensor, and data parallelism. The entire set of devices in a cluster can be allocated in a variety of ways across these three strategies, which can be controlled through the \cd{placement\_strategy} option of the library (see \cite{smp_apidoc}). Regardless of the specific placement, one can define data-parallel group (\dpgroup), pipeline-parallel group (\ppgroup), and tensor-parallel group (\tpgroup) as the sets of processes that collectively perform data parallelism, pipeline parallelism, and tensor parallelism among each other, respectively. In addition, a reduced-data-parallel group (\rdpgroup) is the set of processes that hold the exact same model replica (see Figure~\ref{fig:ranking}). Unlike existing implementations, the library treats {\tpgroup} as a subset of \dpgroup, instead of an independent dimension, which will be explained in more detail in \S\ref{sec:tensor}. We will use \tprank, \pprank, and \dprank to refer to the index of a process within its \tpgroup, \ppgroup, and \dpgroup, respectively. See Figure~\ref{fig:ranking} for an illustration of these concepts.


\subsubsection{User interface}\label{sec:ux}


The core API consists of three main changes to the user script (specific features might require additional APIs, which will be discussed where relevant, see \cite{smp_apidoc} for detailed API documentation, and Appendix~\ref{sec:ap_api} for supplemental documentation). In what follows, we assume that the library is imported through
\begin{verbatim}
import smdistributed.modelparallel.torch \
                    as smp
\end{verbatim} 
To use the library, the user must
\begin{enumerate}
\item Initialize the internal state of the library and launch the backend threads by calling \cd{smp.init()} at the beginning of the script.
\item Wrap the model (\cd{nn.Module} object) with \cd{smp.DistributedModel} wrapper, and optimizer with \cd{smp.DistributedOptimizer} wrapper.
\item Wrap the forward and backward pass logic (but not the optimizer step) with \cd{@smp.step} decorator. For example, the training step might look like
\begin{verbatim}
@smp.step
def train_step(inputs, targets):
    pred = model(inputs)
    loss = loss_obj(pred, targets)
    model.backward(loss)
    return loss, pred
\end{verbatim}
Note that the typical \cd{loss.backward()} call is replaced with \cd{model.backward(loss)} so that the library can control the backward pass. If pipeline parallelism is enabled, \cd{smp.step}-decorated function specifies the computations that must be executed in a pipelined manner. Hence, the computations placed inside the function are executed once per microbatch when \cd{train\_step} function is called. The tensors that are returned from the \cd{smp.step}-decorated function automatically get wrapped in \cd{StepOutput} object, which encapsulates different versions of the tensor across all microbatches. After the call to \cd{train\_step}, these tensors can be combined into a single value using the API exposed by \cd{StepOutput} class (\cd{StepOutput} API available in \cite{smp_apidoc}). For instance, the loss can be averaged, and per-microbatch predictions combined across microbatches through
\begin{verbatim}
# loss is returned by train_step
loss_avg = loss.reduce_mean()
# predictions is returned by train_step
pred = predictions.concat()
\end{verbatim}
After training, to combine model partitions into a single artifact, one can use \cd{state\_dict()} API, which is overriden in \cd{smp.DistributedModel} and \cd{smp.DistributedOptimizer} so that the partitioned model and optimizer states are allgathered and combined in the CPU to produce as a single state dictionary that represents the entire model. It is also possible to get only the local states using \cd{local\_state\_dict()} API, which is useful for checkpointing.
\end{enumerate}

\section{Pipeline Parallelism Architecture}
\label{sec:pipeline}
\subsection{Overview}
Amazon SageMaker model parallelism library views the model as a directed graph of nodes, where each node (with the exception of the root) represents a PyTorch \cd{nn.Module} object, and there is an edge from node $A$ to node $B$ only if module $B$ is a submodule of module $A$ (note PyTorch models are defined as a hierarchy of \cd{nn.Module} objects). Note that the graph is a tree if there are no modules that are submodules of multiple other distinct modules. Module-submodule relationships are derived purely from PyTorch module creations (\cd{\_\_init\_\_} calls), and is independent of how the module is actually used within the \cd{forward} method of the parent module. The root of the graph corresponds to the \cd{smp.step}-decorated function, which is treated as the parent node of the the top-level model object (\cd{smp.DistributedModel}).

\begin{figure}[h]
\centering
\includegraphics[scale=0.45]{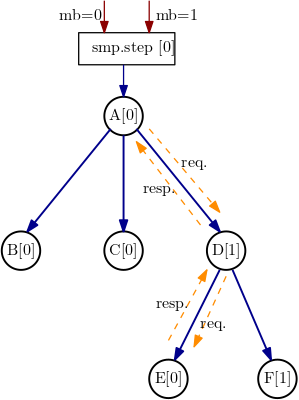}
\caption{The execution flow over an example model with a top-level module $A$, and submodules $B$--$F$. The square brackets represent the {\pprank} the module is assigned to as a result of partitioning. Whenever the control flow reaches a submodule that is not stored locally, an execution request is sent to the {\pprank} that owns the submodule, which responds to the requester with the output of the module. \cd{smp.step} function is always treated as the top module and is placed on {\pprank} 0. For each microbatch, {\pprank} 0 enqueues a new execution request for itself, which marks the start of forward pass. At the start of the backward pass for each microbatch (\cd{model.backward} call), another local request is enqueued at {\pprank} 0, this time for backward execution.}
\label{fig:req_resp}
\end{figure}

The model partition for pipeline parallelism takes place at the level of \cd{nn.Module}, \emph{i.e.}, each partition is a mapping from the set of \cd{nn.Module}s of the model to the set of \pprank{s}. An example module graph and partition is provided in Figure~\ref{fig:req_resp}. Note that there is no assumption on any particular computational graph structure in this view of the model, which is why the pipeline parallelism framework is generic. Further, unlike other implementations of pipeline parallelism, the library does not view the model as a flat sequence of stages (for example, a sequence of layers where certain blocks of layers are assigned to specific devices), but as a \emph{hierarchy} of modules based on the existing hierarchy in model definition. Refer to Appendix~\ref{sec:ap_partition_example} for example partitions over this module hierarchy.

The auto-partitioning algorithm, which assigns modules to \pprank{s}, will be discussed in \S\ref{sec:partition}; for now, we assume the module assignment is given. Every {\pprank} is responsible for (forward and backward) execution of the modules assigned to itself, and stores the necessary parameters for those modules. During forward or backward execution of a module, whenever control flow reaches a submodule that is not assigned to the current {\pprank}, an \emph{execution request} is sent to the {\pprank} that stores that submodule, which in turn sends its own requests to other \pprank{s} if needed. When the execution of the requested module finishes, the result (the outputs of the module for forward pass, or the gradients returned from the backward pass of the module) is returned to the {\pprank} that made the original request.

Each {\pprank} runs an execution server, which is a Python thread that monitors incoming execution requests from other \pprank{s}, and assigns the execution tasks to local worker threads (see \S\ref{sec:moduleserver} for details). During each execution of an \cd{smp.step}-decorated function, instead of following regular Pythonic control flow, every {\pprank} launches the module server and waits for incoming requests. {\pprank} 0 additionally enqueues a new execution request to itself, corresponding to the execution of the top-level \cd{smp.step} node, which consists of the contents of the \cd{smp.step} function. From that point on, execution follows the request-response paradigm. Pipelining across microbatches is achieved by splitting the input tensors to \cd{smp.step}-function across the batch dimension, and initiating microbatch execution requests at different times (See \S\ref{sec:pipeline_pipeline}). At any point in time, requests corresponding to different microbatches are processed by different \pprank{s}, as shown in Figure~\ref{fig:pipeline_schedule}.

Note that this module-server architecture applied over the module hierarchy endows the library with flexibility in handling a range of use cases, such as supporting module reuse, supporting arbitrary model architectures and framework APIs, conditional execution, and handling different pipeline schedules easily, in sharp contrast to other implementations of pipeline parallelism. This is because (1) the module hierarchy is general enough to cover virtually all model implementations in PyTorch (and not those limited to \cd{nn.Sequential} API, or more generally an identical sequence of layers), and (2) the execution at the module servers launched by the \pprank{s} is driven completely dynamically by incoming requests, without any \emph{a priori} expectation about the model architecture or execution flow, (3) auto-partitioning algorithm automatically places modules that share parameters on the same device, (4) re-used modules can be executed simply by sending multiple requests to the device that holds the module.

\begin{figure}[h]
\centering
\includegraphics[scale=0.45]{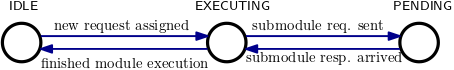}
\caption{The state diagram for a worker thread}
\label{fig:state_diagram}
\end{figure}

\subsection{Module-server execution}\label{sec:moduleserver}
An execution server consists of an input queue, a server for the queue (executed on main thread), a collection of worker threads (Python threads), and a set of modules assigned to the server by the partitioning algorithm. The worker threads do the actual module execution, and can be in one of three states: $\lbp \mathsf{PENDING}, \mathsf{IDLE}, \mathsf{EXECUTING}\rbp$ (see Figure~\ref{fig:state_diagram}). Only one worker thread is allowed to be in $\mathsf{EXECUTING}$ state at a time. Moreover, at any point in time, only one thread per rank (across main thread and worker threads) actively executes, while the rest wait.

\begin{figure}[h]
\centering
\includegraphics[scale=0.32]{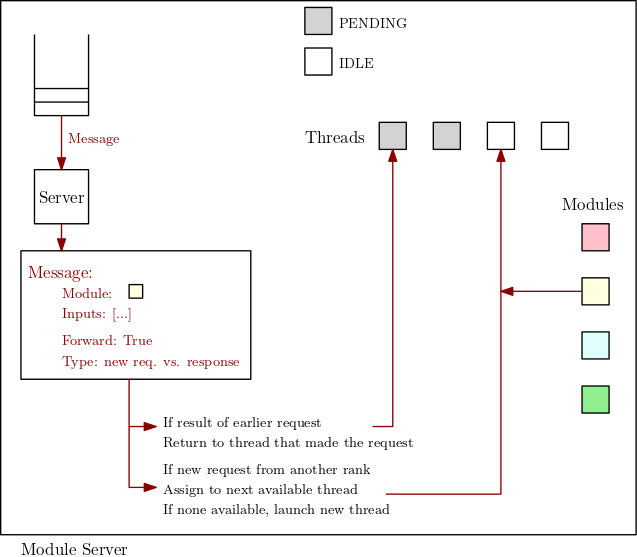}
\caption{The system diagram for a module server}
\label{fig:module_server}
\end{figure}

The server execution works as follows:
\begin{itemize}
\item The tasks requested from an execution server, as well as responses to the requests made by the server, are all enqueued in the input queue. The input queue contains messages that represent the information on what the task is, such as which module needs to be executed, input tensors, whether forward or backward execution is requested, and other metadata (see Appendix~\ref{sec:ap_message} for more details on message structure).
\item When the control is at the main thread, it queries the input queue. When a new request arrives, it looks for the next thread in $\mathsf{IDLE}$  state, and assigns the work to that thread. If no $\mathsf{IDLE}$ thread is available (all threads are in $\mathsf{PENDING}$ state), it launches a new thread, and assigns the module execution task to it. Note that if the control is in main thread, no worker thread can be in $\mathsf{EXECUTING}$ state since only thread can be active at a time. After the worker thread is assigned the work, control switches to it, and it switches to $\mathsf{EXECUTING}$ state.
\item When a worker thread encounters a submodule that is owned by another rank, it sends the request to the appropriate rank, and then switches to $\mathsf{PENDING}$ state, returning control to the main thread. Note that typically multiple threads will be in $\mathsf{PENDING}$ state, each corresponding to a different microbatch.
\item When a response to an earlier request arrives at the queue, it returns the response to the thread that made the request (this thread must be in $\mathsf{PENDING}$ state) and notifies the thread to switch control to it, and switches it into $\mathsf{EXECUTING}$ state.
\end{itemize}
Note that the use of multiple Python threads is purely for the ability to easily context-switch between microbatches, and not for actual parallelization of computation.

\subsection{Pipelining}\label{sec:pipeline_pipeline}
Pipeline schedule is controlled by {\pprank} 0, since it handles the top-level \cd{smp.step}-execution requests. This rank creates two dedicated \cd{smp.step}-execution requests for each microbatch (one for forward, one for backward pass) and assigns it to itself, which marks the start of the forward or backward pass of each microbatch. {\pprank} 0 can determine the pipeline schedule through the timing and sequence of such requests. Note that this rank will be aware of when the forward pass for a particular microbatch finishes through the \cd{model.backward} call.

\begin{figure}[ht]
\includegraphics[scale=0.25]{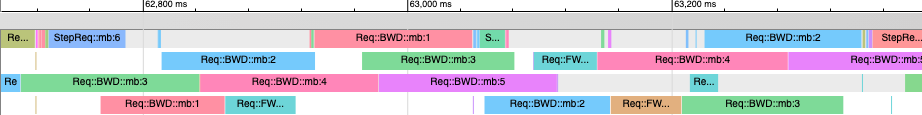}
\caption{Example interleaved pipeline timeline for a 4-billion parameter GPT-2 variant over four \pprank{s}, with 16 microbatches. Processing of forward and backward requests for each microbatch are marked as \cd{Req::FWD} and \cd{Req::BWD} respectively. Note that although backward and forward processing tends to be interleaved, this pattern can be broken opportunistically, depending on the actual timing of events.}
\label{fig:pipeline_schedule}
\end{figure}

The module-server architecture enables supporting custom pipeline schedules flexibly, without making any architectural changes, by simply specifying a sequence of $\lp \cd{microbatch}, \cd{phase}\rp$-tuples, where \cd{phase} can be forward or backward. SageMaker model parallelism library has built-in support for two different pipeline schedules, which are called \emph{simple pipeline} (similar to GPipe), and \emph{interleaved pipeline} (similar to Megatron and PipeDream with pipeline flushes).
\begin{itemize}
\item \textbf{Simple pipeline: } Executes the forward pass of each microbatch sequentially, before executing the backward pass for each microbatch.
\item \textbf{Interleaved pipeline: } If there is a microbatch that is ready to start backward execution, schedules that next. Otherwise, schedules forward execution for the next microbatch. 
\end{itemize}
Note that interleaved pipeline is more flexible and resource-efficient than a rigid, predefined pipeline schedule, since the schedule might adapt opportunistically based on the actual execution times of different modules. See Figure~\ref{fig:pipeline_schedule} for an example pipeline schedule that is measured while training a GPT-2 variant on the module-server architecture, over four partitions.

\subsection{Static mode and fast mode}
Note that when a single module has many submodules that need to be executed sequentially, the module-server architecture might result in unnecessary overhead, since the output of each module needs to be returned to the parent module, which in turn needs to be sent to the next module. Moreover, in most cases, the model and the training step is static, in that the execution flow in every training step is identical. To eliminate the additional overhead in such cases, the library contains two additional features, called \emph{fast mode}, and \emph{static mode}, which can be enabled through flags in model parallelism configuration.

By enabling static mode, the user conveys to the library that the training step does not contain any conditionals, and the flow of execution at every step will be the same. In this case, the library records the sequence of requests/responses at each server for the first $T$ steps (by default, $T=5$), and from $T$th step onward, it avoids explicitly exchanging request and responses, but follows the order of tasks that was followed in the step that achieved the lowest step time among the first $T$ steps. 
In fast mode, which can be enabled on top of static mode, the child $\rightarrow$ parent $\rightarrow$ child round-trip for tensor communication is also avoided, and the tensor directly takes the shortcut from one child module to the next, cutting the amount of communication in half. This is achieved by keeping track of the producer-consumer relationships between modules in the first training step. 

\subsection{Partitioning algorithm}\label{sec:partition}
Partitioning takes place automatically during the first execution of the \cd{smp.step}-decorated function, unless auto-partition is disabled, in which case the user specifies the partition manually using \cd{smp.partition} API. If the model does not fit in the CPU RAM during the initial creation, \cd{smp.delay\_param\_initialization} API can be used to delay parameter allocation until after the model is partitioned and moved to GPUs (see Appendix~\ref{sec:api_delay} for details).

The goal of the partitioning algorithm is to try to minimize the number of request-response communication rounds, while balancing the computation and memory loads across the devices. Intuitively this means that if a node is assigned to a device, its children should be assigned to the same device as much as possible, so that the communication rounds are minimized.

In what follows, we will assume that the model is represented as a tree $\mathcal{T}=\lp \mathcal{V}, \mathcal{E}\rp$ of \cd{ModuleNode}s, which are objects consisting of one or more \cd{nn.Module}s in the model (modules that share parameters are part of the same \cd{ModuleNode}), and the child-parent relationships in the tree follow the hierarchy of modules in the model definition. We define $Q(n)$ as the set of children of a node $n \in \mathcal{V}$: $Q(n) := \lbp j: (n, j) \in \mathcal{E}\rbp$. We assume that for each node $n \in \mathcal{V}$, there is a cost $c(n)$ associated with storing and executing the modules that are part of the subtree of $n$, which accounts for both the memory consumption and computation. The details of how the \cd{ModuleNode} tree and the cost function are constructed is described in Appendix~\ref{sec:ap_partition}. For the purposes of this discussion, we only point out that the cost function satisfies $c(n) \geq \sum_{p \in Q(n)} c(p)$,
\emph{i.e.}, the cost function is super-additive in the subtrees.


We will use the notation $d(n)=i$ to mean that device $i$ was assigned to node $n$. The goal of the algorithm is to compute $d(n)$ for all $n \in \mathcal{T}$.

The algorithm starts with a set of virtual devices $P(r) = \lbp 0, 1, \dots, D-1\rbp$ for the root node $r$ (where $D$ is pipeline parallelism degree specified by user), and operates by traversing the nodes of $\mathcal{T}$ in breadth-first order, and partitioning the set $P(n)$ for the current node $n$ among its children $Q(n)$, so that $P(n) = \bigcup_{c \in Q(n)} P(c)$.

Intuitively, $P(n)$ for a node $n$ represents the set of devices that will be responsible for executing the part of the model represented by the subtree under $n$. At every iteration, the algorithm proceeds by partitioning $P(n)$ among the children of $n$, \emph{i.e.}, deciding which subset of $P(n)$ should be responsible for the execution of each child subtree. For the purposes of the partition algorithm, $P(n)$ refers to ``virtual" devices, in that it pertains to the partition indices within each \ppgroup, and not necessarily physical device indices. The algorithm terminates when $\left| P(n) \right| = 1$ for all the remaining nodes $n$ in the breadth-first traversal, in which case all the remaining children of node $n$ inherit $P(n)$. The current node $n$ always gets assigned the smallest partition index in $P(n)$, denoted by $P(n)[0]$.

\begin{algorithm}[tb]
   \caption{Tree partitioning algorithm}
   \label{alg:tree_partition}
\begin{algorithmic}
   \STATE {\bfseries Input:} Set $P(r)$ of devices, tree $\mathcal{T}$ of nodes with root $r$.
   \WHILE{there are more nodes}
   	\STATE Get next node $n$ in breadth-first order of $\mathcal{T}$
	\STATE $d(n) \leftarrow P(n)[0]$
	\IF{$\left| P(n)\right| > 1$}
		\STATE $\lbp P(c)\rbp_{c \in Q(n)} \leftarrow$ \cd{Partition}$\lp P(n), Q(n)\rp$
	\ELSE
		\STATE $P(c) \leftarrow \lbp P(n)[0] \rbp$ for all $c \in Q(n)$
	\ENDIF
   \ENDWHILE
   \end{algorithmic}
\end{algorithm}


The crux of the algorithm is the choice of how to partition $P(n)$ among the children of node $n$ (\emph{i.e.}, \cd{Partition} call). 
The goal here is to find a partition so that the number of devices assigned to the subtree represented by child $p$ is approximately proportional to the cost of that subtree, $c(p)$, while biasing the algorithm towards assigning the same partition to modules that are adjacent to each other in execution order. The details of how this is done, including a pseudo-code for \cd{Partition} operation (Algorithm~\ref{alg:node_partition}), is given in Appendix~\ref{sec:ap_partition}. The main idea behind this algorithm is to use dynamic programming to split the children $Q(n)$ into segments that are as equal-cost as possible, and then allocating the elements of $P(n)$ across these segments using D'Hondt method \cite{Gallagher_91} (although other proportional allocation methods can be substituted), and recursively re-applying these steps to the segments with more than one device assigned. The intuition behind this is that each child node effectively gets assigned a subset of $P(n)$ that is approximately proportional to the cost of its subtree, so that the per-device cost is balanced. Example partitioning decisions arrived by this algorithm are given in Appendix~\ref{sec:ap_partition_example}.


%
%
%
%
%
%
%

\section{Tensor Parallelism Architecture}
\label{sec:tensor}
\subsection{Motivation}
Tensor parallelism involves splitting operations or layers themselves to execute in parallel across multiple devices. Unlike pipeline parallelism, tensor parallelism needs to be implemented on a per-operation, or per-layer basis, since the distribution mechanism depends on the mathematical function being implemented. For this reason, it is not possible to have a fully model-architecture-agnostic tensor parallelism implementation. However, it is still possible to create a general framework that can efficiently support the full diversity of scenarios that require tensor parallelism (including uniformly large models, models with only one large component, or mixture-of-experts models), and is modular enough to be easily extensible to new custom operations, which are unique advantages of the library compared to other tensor parallelism solutions.

To handle these scenarios efficiently, SageMaker model parallelism library performs tensor parallelism across \emph{data-parallel} ranks, in contrast to other implementations. To see why this matters, consider an example scenario where an otherwise medium-sized model has a huge embedding table that must be distributed across $N$ GPUs. Clearly, using the $N-1$ additional GPUs only for storing the partitions of embedding table is highly inefficient, so these GPUs should contribute to the computation as well. Feeding the same input to all $N-1$ tensor-parallel devices, as Megatron-LM does, would not improve the efficiency in this case, since apart from the embedding, the computation done by the $N-1$ devices would be redundant. It is also impractical to split all the operations in the model since some operations in the model might be difficult to distribute (or unsupported), or might introduce unnecessary cross-device communication for small operations. Performing tensor parallelism across data-parallel ranks effectively solves this problem, by having the GPUs compute on different data samples for the parts of the model that are not distributed, while the relevant parts of the data samples are exchanged between tensor-parallel ranks for the distributed components. We describe the mechanics of tensor parallelism in more detail in the next subsection.

\subsection{Overview}\label{sec:tp_overview}
As with pipeline parallelism, the fundamental computational unit for tensor parallelism is \cd{nn.Module}. In essence, tensor parallelism consists in traversing the model, and replacing specific submodules of the model with their distributed implementations during \cd{smp.DistributedModel} call. The distributed implementation has the same input-output relationship as the original module. A module gets replaced if and only if (1) a distributed implementation of the module is available, (2) the user has enabled tensor parallelism for the module, and (3) no ancestor of the module is already replaced with its distributed version\footnote{In general, distributing a higher-level module is more efficient than distributed multiple lower-level modules, since some of the collectives involved in lower-level distribution can be avoided when distributing a parent module.}, and (4) it does not share parameters with another module.

\begin{figure}[t]
\includegraphics[scale=0.27]{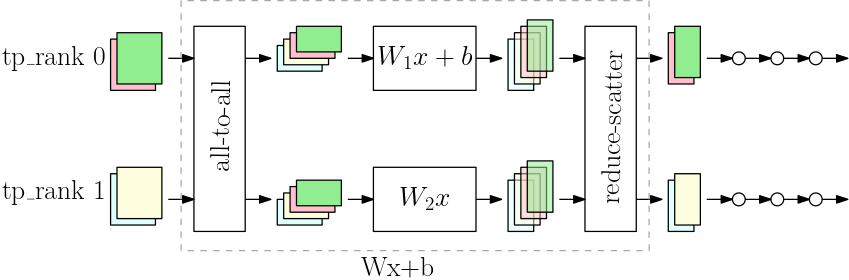}
\caption{Tensor-parallel distribution of an affine layer (\cd{nn.Linear}) over two tensor-parallel ranks. The dashed block implements the affine function $f(x) = Wx+b$, where the weight is distributed column-wise: $W = \lb W_1 \; W_2\rb$ (bias only resides in \tprank{ }0). Each input also gets partitioned row-wise $x = \lb x_1^\top \; x_2^\top \rb^\top$, so that the summation of the two local affine functions: $\lp W_1 x_1 + b\rp + \lp W_2 x_2\rp = Wx + b$ is identical to the non-distributed case. Note that each tensor-parallel rank has a local batch size of 2, consisting of \emph{different} data samples. To implement this for all data samples, \tprank{ }$i$ slices its input $x^{(i)}$ row-wise, and sends the $j$th slice $x_j^{(i)}$ to \tprank{ }$j$ (implemented via all-to-all collective). Next, \tprank{ }$j$ applies its local function to all slices collected from its \tpgroup: $\tilde y_j := \lb y_j^{(1)} \; y_j^{(2)} \rb = W_j \lb x_j^{(1)} \; x_j^{(2)}\rb$ (and adds bias if $j=0$). Finally, a reduce-scatter operation sums $y_j$ across \tprank{s} $j$, and slices it so that \tprank{ }$i$ ends up with $y^{(i)} := \sum_{j=1}^T y_j^{(i)}$, where $T$ is tensor parallelism degree.}
\label{fig:tp_linear}
\end{figure}

When tensor parallelism is performed over data-parallel ranks, the parameters, gradients, and optimizer states for the modules that satisfy (1)--(4) above are partitioned across the \tprank{s}. For the rest of the modules, the tensor-parallel devices operate in a regular data-parallel manner. To execute the distributed module, a device first collects the necessary tensor slices of all data samples across peer devices in the same tensor parallelism group. The local fragment of the module is then executed on the slices of all these data samples, followed by another round of synchronization which both combines the parts of the output for each data sample, and also returns the combined data samples to their respective GPUs where the data sample first originated from, so that the output of the distributed module in every {\tprank} is the same as the non-distributed scenario. We illustrate this idea over an example with \cd{nn.Linear}, depicted in Figure~\ref{fig:tp_linear}. 

The library comes with built-in distributed implementations for commonly used native PyTorch modules, such as \cd{nn.Linear} and \cd{nn.Embedding}, but also has generic implementations for commonly used building blocks in deep learning, such as the attention layer, layer normalization, and transformer encoder/decoder blocks. All tensor-parallel module implementations are child classes of \cd{smp.nn.DistributedModule} class, and part of \cd{smp.nn} module (see Appendix~\ref{sec:api_nn} for APIs for all distributed module implementations).

Note that in order to replace a module with a distributed implementation during \cd{smp.DistributedModel} call, the library needs to be able to match a module with its corresponding \cd{DistributedModule}. For native PyTorch modules, this is achieved using an internal look-up table maintained in the library. This look-up table also includes entries for popular HuggingFace model implementations (GPT-2, RoBERTa, and BERT), which maps specific submodules of these implementations to internal distributed transformer implementations of the library. For custom modules that are not part of the look-up table, \cd{smp.tp\_register} API can be used to register specific \cd{DistributedModule}s with a given module in the user script (See Appendices \ref{sec:ap_tp_api} and ~\ref{sec:api_register}). If there is no built-in distributed implementation for the target module, it is also possible to implement a custom one by sub-classing \cd{smp.nn.DistributedModule}, and using the communication and weight initialization primitives provided the library API, which ensure compatibility with the rest of the features (see Appendix~\ref{sec:api_custom} for details).

\subsection{Built-in distributed modules}
In this section we briefly describe the specific distribution mechanisms implemented for some of the built-in \cd{DistributedModule}s.

{\bf \cd{DistributedLinear}:} The distribution mechanism for the linear layer was described in Figure~\ref{fig:tp_linear}.

{\bf \cd{DistributedEmbedding}:} \cd{nn.Embedding} layers are distributed across the embedding dimension. The input indices are allgathered across the tensor-parallel ranks, followed by a local embedding look-up, which at $i$th {\tprank} gives the $i$th slice of the embedding vector for each index. The results are then scattered across \tprank{s} by the batch dimension, and combined by the embedding dimension, using an all-to-all collective, which gives the local embedding outputs in all \tprank{s}.

{\bf \cd{DistributedTransformer}:} Consists of a sequence of \cd{DistributedTransformerLayer}s, which in turn consist of \cd{DistributedAttentionLayer} and \cd{DistributedTransformerOutputLayer}s. The library offers two separate distribution implementations for the latter two layers, which can be controlled by setting the configuration parameter \cd{"optimize"} to \cd{"speed"} or \cd{"memory"}. The \cd{"speed"} option distributes the attention and output layers in the same way as done by Megatron, while 
\cd{"memory"} option instead shards all the intermediate activations across the \tprank{s}, including layer normalization. See Appendix~\ref{sec:ap_transformer} for details.


\section{Communication Backend}
\label{sec:backend}
The library features a dedicated communication backend for pipeline parallelism, which manages intra-node and cross-node device-to-device (D2D) communication without relying on the popular NCCL library. This is because NCCL necessitates tight synchronization between nodes, requiring collectives (or point-to-point \cd{send}/\cd{recv} operations) be called in the same order in all participating terminals. This is difficult to achieve in the full breadth of cases that the library supports, \emph{e.g.}, conditional control flows where a communication primitive may not be called at a terminal depending on some condition, or cases where two different transmitters simultaneously attempt to send a tensor to the same rank, where a global order of transmissions would require tight synchronization mechanism to be enforced globally.

This motivates a more flexible communication backend, which do not have \emph{a priori} expectations about the order of communications required, and serves communication requests made by the framework in an on-demand basis. Furthermore, for good performance, it needs to leverage NVLinks for intra-node transmissions, and GPUDirect RDMA technology for inter-node transmissions. Appendix~\ref{sec:ap_backend} presents the architecture for the D2D subsystem that satisfies these requirements.

\section{Experiments}
\label{sec:experiments}
\subsection{10-billion parameter RoBERTa and BERT}
We train larger variants of RoBERTa and BERT under two different model configurations shown in Table~\ref{tb:roberta_config}, both totaling 10 billion parameters. These experiments are run on a cluster of 16 \cd{p4d.24xlarge} nodes, each equipped with 8 NVIDIA A100 GPUs.
\begin{table}[t] \small
\caption{Configurations A and B for RoBERTa training. The definitions for model hyperparameters are the same as in \cite{BrownMann_20}}
\label{tb:roberta_config}
\centering
\begin{tabular}{|c|c|c|c|c|}
\hline
{\bf Model} & {\bf $d_{model}$} & {\bf $n_{layers}$} & {\bf $n_{heads}$} & {\bf $d_{head}$} \\
\hline
RoBERTa & 4096 & 48 & 32 & 128 \\
\hline
BERT & 2560 & 127 & 40 & 64 \\
\hline
\end{tabular}
\end{table}
We use a sequence length of 512. For SageMaker, we use a tensor parallelism degree of 4, pipeline parallelism degree of 1, and with activation checkpointing and optimizer state sharding features enabled. For DeepSpeed, we use ZeRO optimization stage 2 with communication overlap and activation (gradient) checkpointing.

The results in Table~\ref{tb:roberta_results} show that for the configuration A, which is a wider architecture, \cd{smp} is 39\% faster than DeepSpeed on 16 nodes. For the deeper configuration B, the two libraries have comparable performance, although DeepSpeed is 15\% faster.
\begin{table}[t] \small
\caption{Throughput of \cd{smp} and DeepSpeed (RoBERTa and BERT).}
\label{tb:roberta_results}
\centering
\begin{tabular}{|c|c|c|c|c|}
\Xhline{3\arrayrulewidth}
{\bf Library} & {\bf Configuration} & {\bf Batch} & {\bf Throughput} \\
\Xhline{3\arrayrulewidth}
\cd{smp} & RoBERTa & 1024 & 385 seq/s \\
\hline
DeepSpeed & RoBERTa & 1024 & 276 seq/s \\
\Xhline{3\arrayrulewidth}
\cd{smp} & BERT & 8192 & 327 seq/s \\
\hline
DeepSpeed & BERT & 8192 & 373 seq/s \\
\Xhline{3\arrayrulewidth}
\end{tabular}
\end{table}


\subsection{175-billion parameter GPT-3}
We train GPT-3 under the 175-billion parameter configuration described in \cite{BrownMann_20}, with a sequence length of 2048. For this experiment, we use 120 \cd{p4d.24xlarge} nodes for a total of 960 NVIDIA A100 GPUs. The library was configured with pipeline parallelism degree 6 (with fast mode enabled), tensor parallelism degree 8, and with activation offloading, activation checkpointing, and optimizer state sharding enabled. To use tensor parallelism with pipeline parallelism, we feed the same data sample to each \tpgroup{ }(see Appendix~\ref{sec:api_config}, section on prescaled batch), so that true data parallelism degree becomes 20. We use a global batch size of 2560, and 64 microbatches. Under this configuration, the library achieves a throughput of 26.5 sequences per second.

\subsection{Neural collaborative filtering with large embedding table}
We train a neural collaborative filtering model \cite{HeLiao_17} with 318133 users, 1792 items, MLP latent dimension of 512, and GMF latent dimension of 64. We use the library with pipeline parallelism degree of 1, and tensor parallelism degree of 16, over four \cd{p3.16xlarge} instances (each with 8 NVIDIA V100 GPUs), and only distribute the four embedding tables of the model, with the rest of the model being executed in a data-parallel manner. We use a per-GPU batch size of 256, for a total batch size of 8192.

As a baseline, we run the same model while feeding the same batch of data to all tensor-parallel ranks, similar to Megatron (note that this removes the need for the initial allgather in \cd{DistributedEmbedding}, as well as turning the all-to-all into an allgather, reducing the amount of communication required). To maintain the same global batch size, we use a per-GPU batch size of 4096. Table~\ref{tb:ncf_results} show the resulting throughput for this model under these configurations for two different internal Amazon datasets.

\begin{table}[ht] \small
\caption{Throughput for \cd{smp} and the baseline over two datasets.}
\label{tb:ncf_results}
\centering
\begin{tabular}{|c|c|c|c|}
\Xhline{3\arrayrulewidth}
{\bf Setting} & {\bf Dataset} & {\bf Throughput} \\
\Xhline{3\arrayrulewidth}
\cd{smp} & 1 & 107656 samples/s \\
\hline
Baseline & 1 & 43078 samples/s  \\
\Xhline{3\arrayrulewidth}
\cd{smp} & 2 & 69298 samples/s \\
\hline
Baseline & 2 &  36723 samples/s \\
\Xhline{3\arrayrulewidth}
\end{tabular}
\end{table}

\bibliography{references}

\begin{thebibliography}{30}
\providecommand{\natexlab}[1]{#1}
\providecommand{\url}[1]{\texttt{#1}}
\expandafter\ifx\csname urlstyle\endcsname\relax
  \providecommand{\doi}[1]{doi: #1}\else
  \providecommand{\doi}{doi: \begingroup \urlstyle{rm}\Url}\fi

\bibitem[Amazon(2020{\natexlab{a}})]{smp_apidoc}
Amazon.
\newblock {Distributed Model Parallelism}.
\newblock
  \url{https://sagemaker.readthedocs.io/en/stable/api/training/smd_model_parallel.html},
  2020{\natexlab{a}}.
\newblock [Online; accessed 29-Sep-2021].

\bibitem[Amazon(2020{\natexlab{b}})]{smp_userguide}
Amazon.
\newblock {Introduction to Model Parallelism - Amazon SageMaker}.
\newblock
  \url{https://docs.aws.amazon.com/sagemaker/latest/dg/model-parallel-intro.html},
  2020{\natexlab{b}}.
\newblock [Online; accessed 29-Sep-2021].

\bibitem[Brown et~al.(2020)Brown, Mann, Ryder, Subbiah, Kaplan, Dhariwal,
  Neelakantan, Shyam, Sastry, Askell, et~al.]{BrownMann_20}
Brown, T.~B., Mann, B., Ryder, N., Subbiah, M., Kaplan, J., Dhariwal, P.,
  Neelakantan, A., Shyam, P., Sastry, G., Askell, A., et~al.
\newblock Language models are few-shot learners.
\newblock \emph{arXiv preprint arXiv:2005.14165}, 2020.

\bibitem[Chen et~al.(2018)Chen, Yang, and Cheng]{ChenYang_18}
Chen, C.-C., Yang, C.-L., and Cheng, H.-Y.
\newblock Efficient and robust parallel dnn training through model parallelism
  on multi-gpu platform.
\newblock \emph{arXiv preprint arXiv:1809.02839}, 2018.

\bibitem[Chen et~al.(2016)Chen, Xu, Zhang, and Guestrin]{ChenXu_16}
Chen, T., Xu, B., Zhang, C., and Guestrin, C.
\newblock Training deep nets with sublinear memory cost.
\newblock \emph{arXiv preprint arXiv:1604.06174}, 2016.

\bibitem[Devlin et~al.(2018)Devlin, Chang, Lee, and Toutanova]{DevlinChang_18}
Devlin, J., Chang, M.-W., Lee, K., and Toutanova, K.
\newblock Bert: Pre-training of deep bidirectional transformers for language
  understanding.
\newblock \emph{arXiv preprint arXiv:1810.04805}, 2018.

\bibitem[Fedus et~al.(2021)Fedus, Zoph, and Shazeer]{FedusZoph_21}
Fedus, W., Zoph, B., and Shazeer, N.
\newblock Switch transformers: Scaling to trillion parameter models with simple
  and efficient sparsity.
\newblock \emph{arXiv preprint arXiv:2101.03961}, 2021.

\bibitem[Gallagher(1991)]{Gallagher_91}
Gallagher, M.
\newblock Proportionality, disproportionality and electoral systems.
\newblock \emph{Electoral studies}, 10\penalty0 (1):\penalty0 33--51, 1991.

\bibitem[Guu et~al.(2020)Guu, Lee, Tung, Pasupat, and Chang]{GuuLee_20}
Guu, K., Lee, K., Tung, Z., Pasupat, P., and Chang, M.-W.
\newblock Realm: Retrieval-augmented language model pre-training.
\newblock \emph{arXiv preprint arXiv:2002.08909}, 2020.

\bibitem[He et~al.(2017)He, Liao, Zhang, Nie, Hu, and Chua]{HeLiao_17}
He, X., Liao, L., Zhang, H., Nie, L., Hu, X., and Chua, T.-S.
\newblock Neural collaborative filtering.
\newblock In \emph{Proceedings of the 26th international conference on world
  wide web}, pp.\  173--182, 2017.

\bibitem[Huang et~al.(2019)Huang, Cheng, Bapna, Firat, Chen, Chen, Lee, Ngiam,
  Le, Wu, et~al.]{HuangCheng_19}
Huang, Y., Cheng, Y., Bapna, A., Firat, O., Chen, D., Chen, M., Lee, H., Ngiam,
  J., Le, Q.~V., Wu, Y., et~al.
\newblock Gpipe: Efficient training of giant neural networks using pipeline
  parallelism.
\newblock \emph{Advances in neural information processing systems},
  32:\penalty0 103--112, 2019.

\bibitem[Lepikhin et~al.(2020)Lepikhin, Lee, Xu, Chen, Firat, Huang, Krikun,
  Shazeer, and Chen]{LepikhinLee_20}
Lepikhin, D., Lee, H., Xu, Y., Chen, D., Firat, O., Huang, Y., Krikun, M.,
  Shazeer, N., and Chen, Z.
\newblock Gshard: Scaling giant models with conditional computation and
  automatic sharding.
\newblock \emph{arXiv preprint arXiv:2006.16668}, 2020.

\bibitem[Lewis et~al.(2020)Lewis, Perez, Piktus, Petroni, Karpukhin, Goyal,
  K{\"u}ttler, Lewis, Yih, Rockt{\"a}schel, et~al.]{LewisPerez_20}
Lewis, P., Perez, E., Piktus, A., Petroni, F., Karpukhin, V., Goyal, N.,
  K{\"u}ttler, H., Lewis, M., Yih, W.-t., Rockt{\"a}schel, T., et~al.
\newblock Retrieval-augmented generation for knowledge-intensive nlp tasks.
\newblock \emph{arXiv preprint arXiv:2005.11401}, 2020.

\bibitem[Li et~al.(2021{\natexlab{a}})Li, Xue, Li, and You]{LiXue_21}
Li, S., Xue, F., Li, Y., and You, Y.
\newblock Sequence parallelism: Making 4d parallelism possible.
\newblock \emph{arXiv preprint arXiv:2105.13120}, 2021{\natexlab{a}}.

\bibitem[Li et~al.(2021{\natexlab{b}})Li, Zhuang, Guo, Zhuo, Zhang, Song, and
  Stoica]{LiZhuang_21}
Li, Z., Zhuang, S., Guo, S., Zhuo, D., Zhang, H., Song, D., and Stoica, I.
\newblock Terapipe: Token-level pipeline parallelism for training large-scale
  language models.
\newblock \emph{arXiv preprint arXiv:2102.07988}, 2021{\natexlab{b}}.

\bibitem[Liu et~al.(2019)Liu, Ott, Goyal, Du, Joshi, Chen, Levy, Lewis,
  Zettlemoyer, and Stoyanov]{LiuOtt_19}
Liu, Y., Ott, M., Goyal, N., Du, J., Joshi, M., Chen, D., Levy, O., Lewis, M.,
  Zettlemoyer, L., and Stoyanov, V.
\newblock Roberta: A robustly optimized bert pretraining approach.
\newblock \emph{arXiv preprint arXiv:1907.11692}, 2019.

\bibitem[Narayanan et~al.(2019)Narayanan, Harlap, Phanishayee, Seshadri,
  Devanur, Ganger, Gibbons, and Zaharia]{NarayananHarlap_19}
Narayanan, D., Harlap, A., Phanishayee, A., Seshadri, V., Devanur, N.~R.,
  Ganger, G.~R., Gibbons, P.~B., and Zaharia, M.
\newblock Pipedream: generalized pipeline parallelism for dnn training.
\newblock In \emph{Proceedings of the 27th ACM Symposium on Operating Systems
  Principles}, pp.\  1--15, 2019.

\bibitem[Narayanan et~al.(2021{\natexlab{a}})Narayanan, Phanishayee, Shi, Chen,
  and Zaharia]{NarayananPhanishayee_21}
Narayanan, D., Phanishayee, A., Shi, K., Chen, X., and Zaharia, M.
\newblock Memory-efficient pipeline-parallel dnn training.
\newblock In \emph{International Conference on Machine Learning}, pp.\
  7937--7947. PMLR, 2021{\natexlab{a}}.

\bibitem[Narayanan et~al.(2021{\natexlab{b}})Narayanan, Shoeybi, Casper,
  LeGresley, Patwary, Korthikanti, Vainbrand, Kashinkunti, Bernauer, Catanzaro,
  et~al.]{NarayananShoeybi_21}
Narayanan, D., Shoeybi, M., Casper, J., LeGresley, P., Patwary, M.,
  Korthikanti, V.~A., Vainbrand, D., Kashinkunti, P., Bernauer, J., Catanzaro,
  B., et~al.
\newblock Efficient large-scale language model training on gpu clusters.
\newblock \emph{arXiv preprint arXiv:2104.04473}, 2021{\natexlab{b}}.

\bibitem[Raffel et~al.(2019)Raffel, Shazeer, Roberts, Lee, Narang, Matena,
  Zhou, Li, and Liu]{RaffelShazeer_19}
Raffel, C., Shazeer, N., Roberts, A., Lee, K., Narang, S., Matena, M., Zhou,
  Y., Li, W., and Liu, P.~J.
\newblock Exploring the limits of transfer learning with a unified text-to-text
  transformer.
\newblock \emph{arXiv preprint arXiv:1910.10683}, 2019.

\bibitem[Rajbhandari et~al.(2020)Rajbhandari, Rasley, Ruwase, and
  He]{RajbhandariRasley_20}
Rajbhandari, S., Rasley, J., Ruwase, O., and He, Y.
\newblock Zero: Memory optimizations toward training trillion parameter models.
\newblock In \emph{SC20: International Conference for High Performance
  Computing, Networking, Storage and Analysis}, pp.\  1--16. IEEE, 2020.

\bibitem[Rajbhandari et~al.(2021)Rajbhandari, Ruwase, Rasley, Smith, and
  He]{RajbhandariRuwase_21}
Rajbhandari, S., Ruwase, O., Rasley, J., Smith, S., and He, Y.
\newblock Zero-infinity: Breaking the gpu memory wall for extreme scale deep
  learning.
\newblock \emph{arXiv preprint arXiv:2104.07857}, 2021.

\bibitem[Rasley et~al.(2020)Rasley, Rajbhandari, Ruwase, and
  He]{RasleyRajbhandari_20}
Rasley, J., Rajbhandari, S., Ruwase, O., and He, Y.
\newblock Deepspeed: System optimizations enable training deep learning models
  with over 100 billion parameters.
\newblock In \emph{Proceedings of the 26th ACM SIGKDD International Conference
  on Knowledge Discovery \& Data Mining}, pp.\  3505--3506, 2020.

\bibitem[Real et~al.(2019)Real, Aggarwal, Huang, and Le]{RealAggarwal_19}
Real, E., Aggarwal, A., Huang, Y., and Le, Q.~V.
\newblock Regularized evolution for image classifier architecture search.
\newblock In \emph{Proceedings of the aaai conference on artificial
  intelligence}, volume~33, pp.\  4780--4789, 2019.

\bibitem[Ren et~al.(2021)Ren, Rajbhandari, Aminabadi, Ruwase, Yang, Zhang, Li,
  and He]{RenRajbhandari_21}
Ren, J., Rajbhandari, S., Aminabadi, R.~Y., Ruwase, O., Yang, S., Zhang, M.,
  Li, D., and He, Y.
\newblock Zero-offload: Democratizing billion-scale model training.
\newblock \emph{arXiv preprint arXiv:2101.06840}, 2021.

\bibitem[Shazeer et~al.(2017)Shazeer, Mirhoseini, Maziarz, Davis, Le, Hinton,
  and Dean]{ShazeerMirhoseini_17}
Shazeer, N., Mirhoseini, A., Maziarz, K., Davis, A., Le, Q., Hinton, G., and
  Dean, J.
\newblock Outrageously large neural networks: The sparsely-gated
  mixture-of-experts layer.
\newblock \emph{arXiv preprint arXiv:1701.06538}, 2017.

\bibitem[Shazeer et~al.(2018)Shazeer, Cheng, Parmar, Tran, Vaswani,
  Koanantakool, Hawkins, Lee, Hong, Young, et~al.]{ShazeerCheng_18}
Shazeer, N., Cheng, Y., Parmar, N., Tran, D., Vaswani, A., Koanantakool, P.,
  Hawkins, P., Lee, H., Hong, M., Young, C., et~al.
\newblock Mesh-tensorflow: Deep learning for supercomputers.
\newblock \emph{arXiv preprint arXiv:1811.02084}, 2018.

\bibitem[Shoeybi et~al.(2019)Shoeybi, Patwary, Puri, LeGresley, Casper, and
  Catanzaro]{ShoeybiPatwary_19}
Shoeybi, M., Patwary, M., Puri, R., LeGresley, P., Casper, J., and Catanzaro,
  B.
\newblock Megatron-lm: Training multi-billion parameter language models using
  model parallelism.
\newblock \emph{arXiv preprint arXiv:1909.08053}, 2019.

\bibitem[Thangakrishnan et~al.(2020)Thangakrishnan, Cavdar, Karakus, Ghai,
  Selivonchyk, and Pruce]{ThangakrishnanCavdar_20}
Thangakrishnan, I., Cavdar, D., Karakus, C., Ghai, P., Selivonchyk, Y., and
  Pruce, C.
\newblock Herring: rethinking the parameter server at scale for the cloud.
\newblock In \emph{SC20: International Conference for High Performance
  Computing, Networking, Storage and Analysis}, pp.\  1--13. IEEE, 2020.

\bibitem[Xu et~al.(2021)Xu, Lee, Chen, Hechtman, Huang, Joshi, Krikun,
  Lepikhin, Ly, Maggioni, et~al.]{XuLee_21}
Xu, Y., Lee, H., Chen, D., Hechtman, B., Huang, Y., Joshi, R., Krikun, M.,
  Lepikhin, D., Ly, A., Maggioni, M., et~al.
\newblock Gspmd: General and scalable parallelization for ml computation
  graphs.
\newblock \emph{arXiv preprint arXiv:2105.04663}, 2021.

\end{thebibliography}
\bibliographystyle{mlsys2022}

\clearpage
\newpage

\appendix
\section{Model Parallelism Basics}
\label{sec:ap_modelparallel}
Model parallelism is a type of distributed training pattern where a single copy of the model is partitioned across multiple accelerators which operate in parallel. In contrast to data parallelism, which is useful at all model scales, model parallelism becomes more and more useful at larger model scales, and becomes unavoidable beyond a certain scale, absent some other powerful memory-saving technique, such as tensor sharding across data-parallel ranks, or CPU offloading\footnote{Often such techniques will be used in combination with model parallelism}.

There are two main types of model parallelism: (1) pipeline parallelism, and (2) tensor parallelism. SageMaker model parallelism library supports both types of model parallelism.

\begin{figure}[ht]
\centering
\includegraphics[scale=0.2]{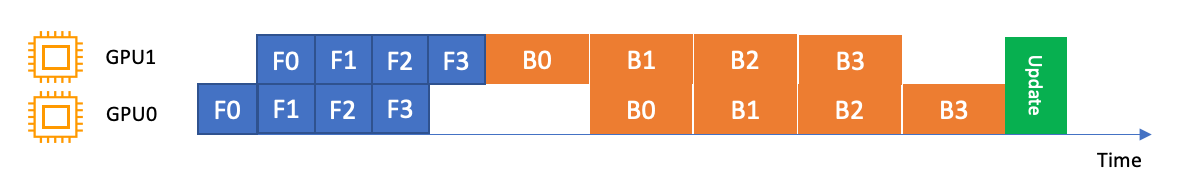}
\caption{Example simple pipeline schedule over 2 GPUs}
\label{fig:simple}
\end{figure}

\begin{figure}[ht]
\centering
\includegraphics[scale=0.25]{interleaved.png}
\caption{Example interleaved pipeline schedule over 2 GPUs}
\label{fig:interleaved}
\end{figure}

\subsection{Pipeline parallelism}
Pipeline parallelism partitions the set of layers or modules, or, depending on the granularity of the partition, operations across the set of devices, such that each operation remains fully within one device. The pipeline-parallel partition is often represented as a sequence of stages in the model, \emph{e.g.}, the first $N/2$ layers of the model is stored in one device, and the next $N/2$ is stored on another etc.

Note that since the partition typically consists of a set of sequential stages that depend on each other, such partitions do not immediately achieve parallelization. To improve parallelization, \emph{pipelined execution} is performed, where the incoming batch of data is split into \emph{microbatches}, which are simply subsets of a single mini-batch. Each microbatch is sequentially fed into the model, and follows a tightly orchestrated pipeline schedule that prescribes which stage of computation (forward or backward) for which microbatch each device should perform for each time slot. The pipeline schedule is typically designed so that the amount of parallelization (simultaneous computation) between the devices is maximized (see Figures~\ref{fig:simple} and \ref{fig:interleaved}). Works such as GPipe, PipeDream, Megatron, and DeepSpeed implement different variants of pipeline parallelism in limited ways (See Section~\ref{sec:relatedwork} for a brief comparison of these works).

\subsection{Tensor parallelism}
In contrast to pipeline parallelism, tensor parallelism consists of splitting specific operations and implementing them in a distributed way. For instance, pipeline parallelism would leave each matrix multiplication operation intact, while tensor parallelism partitions the matrices to implement distributed matrix multiplication. Unlike pipeline parallelism, tensor parallelism results in immediately parallelized computations.

Note that the specific way in which an operation is distributed in tensor parallelism depends on the mathematical function implemented by the operation (\emph{e.g.}, distributed matrix multiplication is implemented differently from distributed embedding look-up), and thus tensor parallelism must be implemented in an operator-specific basis.

Works such as Megatron \cite{ShoeybiPatwary_19}, Mesh-TensorFlow, and GSPMD implement variants of tensor parallelism.

\section{Pipeline parallelism message structure}
\label{sec:ap_message}
The messages exchanged between \pprank{s} contain the following information:
\begin{itemize}
\item Whether this is a request or response, and if it is a response, an identifier for the corresponding request
\item If request, an identifier for the module the execution is requested for
\item If a request for a \cd{nn.Sequential} module, the subset of (consecutive) modules the execution is requested for
\item If a request for module execution, whether forward or backward execution is requested
\item If a request for module execution, the input tensors for the module (along with any encapsulating Python structure such as \cd{list}s that contain the tensors) or the gradients flowing into the module
\item If a response, module outputs (along with any encapsulating Python structure such as \cd{list}s that contain the tensors) or the gradients flowing out of the module
\item Microbatch information
\item The \pprank making the request
\item The location of the module within the module graph
\item Whether the requester is currently in a \cd{autocast} context
\item Whether the requester is currently in a \cd{no\_grad}/\cd{enable\_grad} context
\item Whether activation checkpointing is enabled for the current module
\end{itemize}

\section{Partitioning algorithm details}
\label{sec:ap_partition}
\subsection{Tracing} 
During the first execution of the \cd{smp.step}-decorated function, before the true forward pass starts, the library first runs a tracing step, which consists of a single forward pass execution of the model on \rank 0. The goal of the tracing step is to determine the order in which the modules are executed, as well as profiling metrics such as the execution time of each module and activation memory consumption, which are fed as inputs to the partitioning algorithm. 

The only significant memory usage during the tracing step is the model parameters (activations, gradients, and optimizer states are not stored), so tracing can often be done on a single GPU up to a certain model size (approximately 15-20 billion parameters). For larger models, the library also supports tracing on CPU (with retries on GPU for modules unsupported on CPU), which typically has access to much larger memory. In this case, module execution times are not measured.

Note that for larger models, even a single CPU memory is not sufficient. In those cases, we do not perform tracing.

\subsection{Tree construction}
The partitioning algorithm operates on a tree of \cd{ModuleNode}s, where each \cd{ModuleNode} consists of one or more \cd{nn.Module}s, and each \cd{nn.Module} is covered by exactly one \cd{ModuleNode}. The mapping from \cd{nn.Module}s to \cd{ModuleNode}s are as follows. 

Define the (bipartite) graph $\mathcal{G}$ which has the set of \cd{nn.Module}s and \cd{nn.Parameter}s as its vertices, and there is an edge between two vertices $i$ and $j$ if and only if module $i$ contains parameter $j$. Then each connected component in $\mathcal{G}$ represents a \cd{ModuleNode}. Intuitively, a \cd{ModuleNode} consists of a set of modules that share a set of parameters. In practice, almost all \cd{ModuleNode}s will contain only a single module, since most modules for models in practice do not share parameters.

Given such \cd{ModuleNode}s, a tree can be constructed by adding an edge from $i$ to $j$ if $i$ contains a module that has a module in $j$ as its child, and in case of multiple parents, pruning the additional edges so that a tree is produced in the end. Since the pruning is arbitrary, the trees that can be produced by this process is not unique, but we simply choose one such tree, with the assumption that most modules do not share their parameters, hence the resulting trees are mostly similar to each other.


\subsection{Cost function}
We assign a cost to each module as follows:
\begin{align}
\bar C(m) := \alpha w(m) + (1 - \alpha) \psi (m),
\end{align}
where $\bar C(m)$ is the unnormalized cost, $w(m)$ is the memory cost, $\psi(m)$ is the computation cost of module $m$; and $\alpha \in \lb 0, 1 \rb$ is a hyperparameter that governs the trade-off between balancing the memory and balancing the computation. If module execution times are measured during tracing, $\psi(m)$ is the forward execution time of module $m$; otherwise, it is the number of descendant modules it contains, which is treated as a proxy for the computational load of the module. $w(m)$ is computed through the number of parameters contained in module $m$ including its submodules, plus the activation memory usage measured during tracing.

The cost $C(n)$ of a \cd{ModuleNode} $n$ is recursively defined as the sum of the costs of the set of modules $M(n)$ it contains and the costs of its children $Q(n)$:
\begin{align*}
C(n) = \sum_{m \in M(n)} \bar C(m) + \sum_{p \in Q(n)} C(p).
\end{align*}
Finally, the normalized cost $c(n)$ of a \cd{ModuleNode} $n$ is defined as the cost of $n$ divided by the cost of the root $r$:
\begin{align*}
c(n) = \frac{C(n)}{C(r)},
\end{align*}
where $C(r) > 0$ since there is at least one module and one parameter in the model. Note that $c(n) \in \lb 0, 1\rb$, and $c(r) = 1$.

\subsection{Partitioning algorithm}

\begin{algorithm}[tb]
   \caption{\cd{Partition}$\lp P, \lbp n_i \rbp_{0 \leq i \leq m-1}\rp$}
   \label{alg:node_partition}
\begin{algorithmic}
   \STATE {\bfseries Input:} Set $P$ of devices, nodes $n_j$, for $0 \leq j \leq m-1$.
   	\STATE Find $\ell$ segments $\lbp S_i \rbp_{0 \leq i \leq \ell-1}$ by solving \eqref{eq:dp_opt} using the recursion \eqref{eq:dp_recursion}.
	\STATE Compute segment costs $c_i := \sum_{n \in S_i} c(n)$, $0 \leq i \leq \ell-1$
	\STATE Compute segment allocations \\ \;\;\;\; $\lbp P_i\rbp_{0 \leq i \leq \ell - 1} \leftarrow$ Algorithm~\ref{alg:dhondt} $\lp P, \lbp c_i\rbp_{0 \leq 0 \leq \ell-1}\rp$
	\FOR{$i=0$ {\bfseries to} $\ell-1$}
		\IF{$\left| P_i\right| == 0$}
			\STATE For all nodes $n \in S_i$, set $P(n) = \lbp P[0] \rbp$
		\ELSIF{$\left| S_i \right| == 1$ {\bfseries or} $\left| P_i \right| == 1$}
			\STATE Set $P(n) = P_i$ for all nodes $n \in S_i$
		\ELSE
				\STATE Run \cd{Partition$\lp P_i, \lbp n_j \rbp_{j \in S_i}\rp$}
		\ENDIF
	\ENDFOR
	\STATE {\bfseries return } $\lbp P(n_i)\rbp$ for $0 \leq i \leq m-1$
   \end{algorithmic}
\end{algorithm}

\begin{algorithm}[tb]
   \caption{D'Hondt method}
   \label{alg:dhondt}
\begin{algorithmic}
   \STATE {\bfseries Input:} Set $P(n)$ of devices, and costs $c_i > 0$, for $0 \leq i \leq \ell-1$.
   \STATE Initialize s:=1, $q_i := c_i$, $P_i = \lbp \rbp$ for $0 \leq i \leq \ell-1$.
   \FOR{$p\in P(n)$}
	\STATE $P_k \leftarrow P_k \cup \lbp p\rbp$ where $k$ := $\arg\max_i q_i$
	\STATE $q_k \leftarrow \frac{q_k}{s+1}$
	\STATE $s \leftarrow s+1$
   \ENDFOR
   \STATE {\bfseries return } $\lbp P_i\rbp$ for $0 \leq i \leq \ell-1$
   \end{algorithmic}
\end{algorithm}

In this section, we describe how we partition the set $P(n)$ of devices among the children $Q(n)$ of node $n$, so that
\begin{align*}
P(n) = \bigcup_{c \in Q(n)} P(c).
\end{align*}
Note that the sets $P(c)$ may not be disjoint across $c \in Q(n)$, and it is possible that $\left|P(c)\right|>1$, in which case $P(c)$ will further be partitioned among the children of $c$.

To achieve this, we first partition the sequence of $Q(n)$ (where the order is dictated by the execution order obtained during tracing) into $\ell$ segments such that maximum normalized cost in any segment is minimized:
\begin{align}
\min_{\mathcal{P}} \omega(\mathcal{P}) := \min_{\mathcal{P}} \max_{S \in \mathcal{P}} \sum_{p \in S} c(p), \label{eq:dp_opt}
\end{align}
where $\mathcal{P}$ is a partitioning of the sequence $Q(n)$ into $\ell$ subsequences $S$ with consecutive elements. In other words, we are seeking an optimally balanced $\ell$-way partitioning of $Q(n)$ into sub-arrays with consecutive elements. This can be solved through dynamic programming, using the recursion
\begin{align}
c(k, i) = \min_{j \leq i} \max \lbp c(k-1, j), \sum_{q \in Q(n, j)} c(q)\rbp,  \label{eq:dp_recursion}
\end{align}
for $0 \leq i < \left| Q(n)\right|$, $2 \leq k \leq \ell$, where $c(k,i)$ is defined as the partition cost $\omega(\mathcal{P})$ achieved in partitioning the first $i$ elements of $Q(n)$ into $k$ partitions, and $Q(n, j)$ represents the sub-sequence of $Q(n)$ from element $j$ onwards. To see how recursion \eqref{eq:dp_recursion} solves \eqref{eq:dp_opt}, note that $\min_\mathcal{P} \omega(\mathcal{P}) = c(\left| Q(n)\right|, \ell)$, and $c(1, i) = \sum_{j \leq i} c(j)$. In practice, we choose $\ell$ to be node-dependent: $\ell = \left|P(n)\right|$.

Once the $\ell$ segments are formed, we allocate the virtual devices $P(n)$ across the $\ell$ segments, such that the number of devices assigned to segment $S$ is approximately proportional to the total cost $\sum_{p \in S} c(p)$ of segment $i$. Any proportional allocation algorithm can be used to achieve this, but we implement D'Hondt method\footnote{It can be shown that D'Hondt method minimizes the largest device-count-to-segment-cost ratio among all segments. The method is used in the allocation of parliament seats proportional to the votes in many representative democracies around the world. Note the similarity of the parliament seat allocation problem with the virtual device allocation proportional to segment costs.} \cite{Gallagher_91} in practice to perform this allocation since it tends to favor larger-cost segments (the motivation for this will become clear shortly). D'Hondt algorithm is outlined in Algorithm~\ref{alg:dhondt}.

\begin{figure*}[t!]
\centering
\includegraphics[scale=0.45]{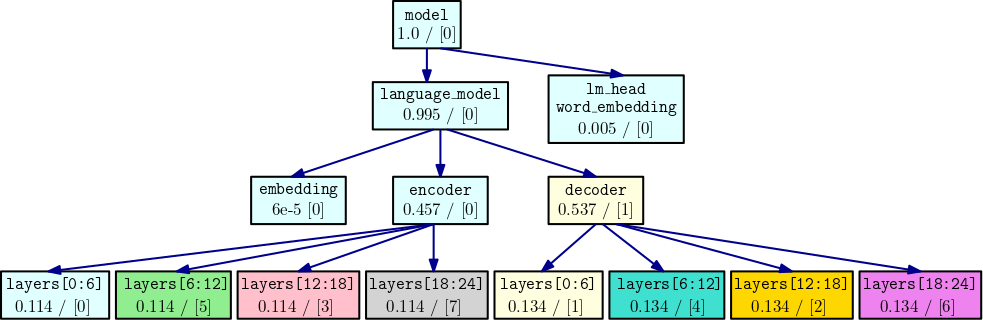}
\caption{Auto-partition decision over T5-11B with $\alpha=1.0$. \cd{layers[i:j]} represent encoder/decoder layers between $i$ (inclusive) and $j$ (exclusive). The numbers on the second row of each box represent the normalized cost of the corresponding \cd{ModuleNode}, and the partition index the node is assigned to, respectively. Note that since \cd{lm\_head} and \cd{word\_embedding} share weights, they are assigned to the same \cd{ModuleNode}, and are thus assigned to the same partition. Per-partition total normalized costs vary between 0.114 and 0.134 (note that perfect balancing would assign 0.125 to each partition).}
\label{fig:t5_10}
\end{figure*}

At the end of D'Hondt allocation, there are three possible scenarios for each segment:
\begin{enumerate}
\item No device is assigned to the segment. In this case, all \cd{ModuleNode}s in the segment get assigned the same device as the parent node\footnote{This can happen when the cost of the segment is too small, in which case assigning the same virtual device as the parent avoids an additional round of communication. D'Hondt method biases towards this scenario by slightly favoring the larger-cost segments in its allocation.}.
\item One device is assigned to the segment. In this case, all \cd{ModuleNode}s in the segment get assigned this device.
\item Multiple devices are assigned to the segment. In this case, if the segment has one node, we keep the device assignment as is (this node will be revisited later in BFS). Otherwise, we recursively apply the dynamic programming and D'Hondt allocation steps to this segment, until each sub-segment reduces to one of the above two scenarios. Note that this process is guaranteed to terminate as long as all \cd{ModuleNode} costs are strictly positive.
\end{enumerate}

To understand the intuition behind the algorithm, it is useful to consider how it would behave in some specific scenarios. Note that Algorithm~\ref{alg:node_partition} must be versatile enough to handle diverse node cost distributions. For instance, the children nodes $Q(n)$ might consist of one or two nodes that contain most of the total cost, along with a large number of tiny-cost nodes. On the other extreme, the children might consist of a large number of nodes for which the cost is almost-equally distributed. In the former scenario, the segmentation step will combine all the small nodes into a single segment, and the large-cost nodes will be placed in their dedicated segments. If the combined cost of the combined nodes is small enough, D'Hondt method will allocate all the devices to the nodes that account for most of the cost, while the small-cost nodes will simply be placed on the same device as the parent, which is desirable. In the latter scenario, segmentation will create approximately-equal-cost segments, and D'Hondt method will allocate one device for each segment (recall that we choose the number of segments to be equal to the number of devices for the current node), balancing the per-device load. In practice, we observe that the presented algorithm can handle a broad range of scenarios across different model architectures and implementations, and result in a relatively balanced partition, although manual tuning might somtimes improve the balance of the memory load (See \cite{smp_apidoc} for the manual partition API).

\section{Example partition decisions}
\label{sec:ap_partition_example}
\begin{figure*}[t!]
\centering
\includegraphics[scale=0.45]{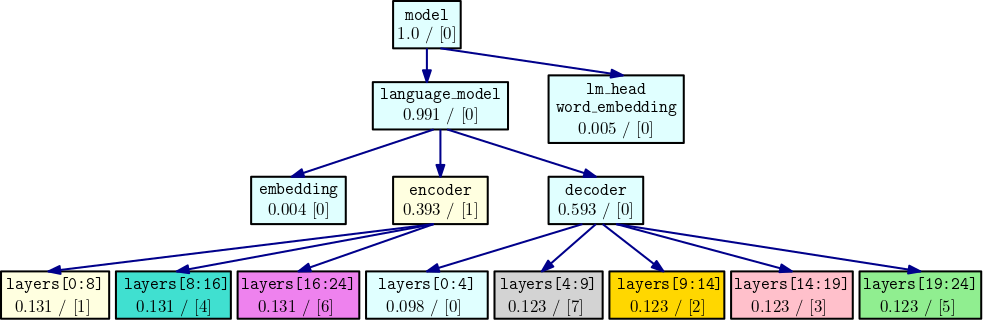}
\caption{Auto-partition decision over T5-11B with $\alpha=0.2$. See Figure~\ref{fig:t5_10} for an explanation of the representation. The encoder gets assigned three devices while the decoder is assigned five, since decoder involves more computation, and setting $\alpha=0.2$ results in weighing the computational component higher in computing the normalized cost. Per-partition total normalized costs vary between 0.107 and 0.131 (note that perfect balancing would assign 0.125 to each partition).}
\label{fig:t5_02}
\end{figure*}

Figures~\ref{fig:t5_10} and \ref{fig:t5_02} illustrate partition decisions output by the auto-partitioning algorithm over a T5 model with 11 billion parameters, with memory-cost weights $\alpha=1.0$ and $\alpha=0.2$, respectively, and a pipeline parallelism degree of 8.

In either case, the partitioning algorithm balances the normalized costs across 8 partitions. In the case of $\alpha=1.0$, note that the costs of encoder and decoder blocks are closer, and so the auto-partitioning algorithm assigns four devices to each. In contrast, for $\alpha=0.2$, the decoder block has a relatively higher cost, and hence gets assigned five devices, compared to three for the encoder. The additional cost of the decoder reflects the fact that it performs the additional cross-attention operation.

Further note that in Figure~\ref{fig:t5_02}, partition 0 gets assigned fewer decoder layers, balancing the fact that it holds other layers such as embedding and \cd{lm\_head}. 

\section{Tensor parallelism API}
\label{sec:ap_tp_api}
\subsection{\cd{smp.nn} module}
All distributed module implementations inherit from \cd{smp.nn.DistributedModule}. \cd{smp.nn} module contains all the built-in implementations of distributed modules, which are as follows:
\begin{itemize}\setlength\itemsep{0em}
\item \cd{smp.nn.DistributedLinear}
\item \cd{smp.nn.DistributedEmbedding}
\item \cd{smp.nn.DistributedLayerNorm}
\item \cd{smp.nn.DistributedAttentionLayer}
\item \cd{smp.nn.DistributedTransformerOutputLayer}
\item \cd{smp.nn.DistributedTransformerLayer}
\item \cd{smp.nn.DistributedTransformer}
\item \cd{smp.nn.DistributedTransformerLMHead}
\end{itemize}
These distributed implementations are built in a generic manner, supporting a range of use cases across different model architectures through initialization arguments, such as self- vs. cross-attention for encoder-decoder architectures, causal vs. non-causal masks for language modeling (\emph{e.g.}, BERT vs. GPT-2), and pre- vs. post-residual layer normalization in transformer layers. More details on the API documentation for these distributed modules can be found in Appendix~\ref{sec:api_nn}.

\subsection{Enabling tensor parallelism}
Tensor parallelism can be used by either using the distributed modules listed in the previous section directly in model construction phase, or the library can automatically replace the modules with their distributed implementations whenever possible, as requested by the user. Such automated replacement can be useful, for instance, when the user do not have direct access to the model construction code (such as when importing a HuggingFace transformer implementation), or many sub-modules in the model require distribution.

For modules whose distribution is natively supported in the library, such automatic replacement can be enabled by
\begin{verbatim}
with smp.tensor_parallelism(enabled=True):
    module = MyModule()
\end{verbatim}
which marks \cd{MyModule}, as well as all of its submodules, for tensor parallelism. Later, when \cd{smp.DistributedModel} wrapper is called, all modules marked for tensor parallelism are replaced with their distributed implementation, as long as they satisfy the conditions (1)--(4) listed in Section~\ref{sec:tp_overview}. Alternatively, tensor parallelism can be enabled for a specific module using the \cd{smp.set\_tensor\_parallelism} API, for example,
\begin{verbatim}
smp.set_tensor_parallelism(model.embedding, 
        enabled=True)
\end{verbatim}

Modules with built-in tensor parallelism support include \cd{nn.Linear} and \cd{nn.Embedding} from native PyTorch, as well as the relevant submodules of HuggingFace \cd{transformers} implementations for BERT, RoBERTa, and GPT-2.

\subsection{Registering a distributed module for tensor parallelism}
For custom module implementations without built-in support, the user can register the module class with a built-in \cd{DistributedModule} using \cd{@smp.tp\_register} decorator or \cd{smp.tp\_register\_with\_module} API, which informs the library that the module implements the same function as the corresponding \cd{DistributedModule}. During registration, the user may specify hooks which translate the \cd{\_\_init\_\_} and \cd{forward} arguments and return values from the original module to the distributed one:
\begin{verbatim}
@smp.tp_register(
    smp.nn.DistributedAttentionLayer, 
    init_hook, fwd_hook)
class CustomAttentionLayer(nn.Module):
    ...
\end{verbatim}
or
\begin{verbatim}
smp.tp_register_with_module(
    CustomAttentionLayer, 
    smp.nn.DistributedAttentionLayer)
\end{verbatim}
Once registered, if tensor parallelism is enabled for the original module, the library replaces the module with its registered distributed counterpart, and matches its hyperparameters and method signature using the hooks defined during registration. Note that such registration is not needed if the user directly imports and uses the \cd{DistributedModule} during model construction. The details for these APIs are available in Appendix~\ref{sec:api_register}.

\subsection{Creating custom distributed modules}
For modules whose distributed implementation is not available among built-in modules listed in Section~\ref{sec:api_nn}, the user can also implement custom distributed modules by sub-classing \cd{smp.nn.DistributedModule}, and registering them with the existing modules. The library API exposes a number of primitives aiding with distributed weight initialization and collective communication, which internally handle the interactions with other features such as data parallelism and pipeline parallelism, so that new custom distributed modules can be implemented easily. These primitives are described in detail in Appendix~\ref{sec:api_custom}.

\section{Distributed Transformer Implementations}
\label{sec:ap_transformer}

\begin{figure}[t]
\centering
\includegraphics[scale=0.3]{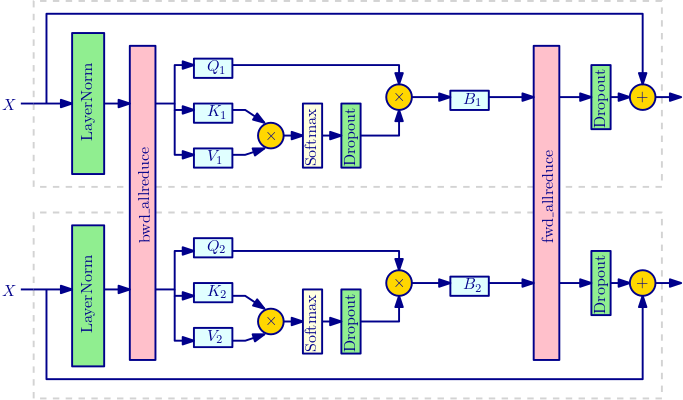}
\caption{\cd{DistributedTransformerOutputLayer} implementation when \cd{optimize}: \cd{"speed"}. \cd{bwd\_allreduce} operation is implemented by the \cd{bwd\_allreduce\_for\_tp} collective (see Appendix~\ref{sec:api_custom})}.
\label{fig:attention_speed}
\end{figure}

In this section we present the distribution mechanisms for the self-attention and MLP blocks of a transformer. The library features two separate distribution mechanisms for optimizing memory footprint, and throughput, respectively. The former avoids replicating activations within the \tpgroup, while the latter minimizes communication across \tprank{s}, and is effectively equivalent to the distribution method of Megatron-LM. The distribution method can be chosen by setting \cd{"optimize": "speed"} or \cd{"optimize": "memory"} in the configuration while launching the job.

\subsection{Speed-optimized distribution}
\begin{figure}[t]
\centering
\includegraphics[scale=0.3]{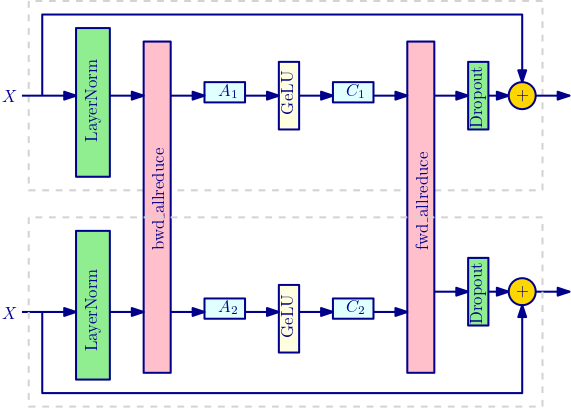}
\caption{\cd{DistributedTransformerOutputLayer} implementation when \cd{optimize}: \cd{"speed"}. \cd{bwd\_allreduce} operation is implemented by the \cd{bwd\_allreduce\_for\_tp} collective (see Appendix~\ref{sec:api_custom})}
\label{fig:mlp_speed}
\end{figure}

In speed-optimized distribution, within each transformer layer, the first linear layers are distributed across their output channels, and the second ones are distributed across their input channels, for both self-attention, and MLP blocks. The activations sharded between these two blocks, and are replicated across \tprank{s} otherwise. To execute a single transformer layer, two allreduces during forward, and two allreduces during backward pass are required. The system diagram for self-attention and MLP blocks are provided in Figures~\ref{fig:attention_speed} and \ref{fig:mlp_speed}. Note that this distribution mechanism is the same as that implemented by Megatron-LM \cite{ShoeybiPatwary_19}.

\subsection{Memory-optimized distribution}
Memory-optimized distribution offers a novel method for tensor-parallel distribution of the transformer layer, where all linear layers are distributed over their input channels, and are followed by a reduce-scatter operation in forward pass, which sums the tensors and slices the result across the channel dimension ($i$th {\tprank} ends up with the $i$th slice of the summed tensor). The reduce-scatter operation becomes an allgather in the backward pass.

\begin{figure}[t]
\centering
\includegraphics[scale=0.3]{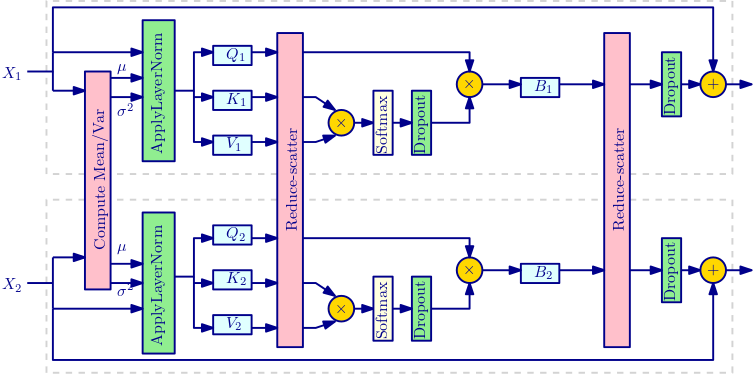}
\caption{\cd{DistributedAttentionLayer} implementation when \cd{optimize}: \cd{"memory"}}
\label{fig:attention_memory}
\end{figure}

\begin{figure}[t]
\centering
\includegraphics[scale=0.3]{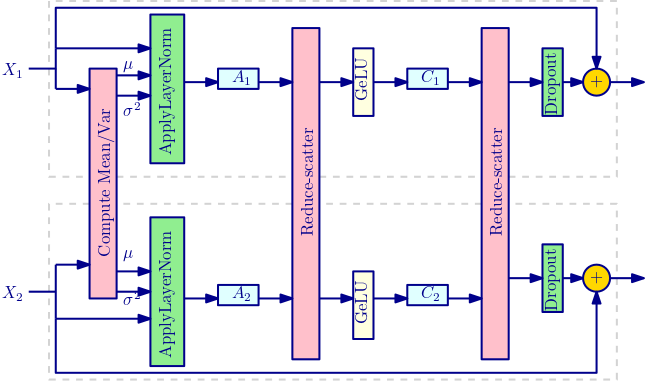}
\caption{\cd{DistributedTransformerOutputLayer} implementation when \cd{optimize}: \cd{"memory"}}
\label{fig:mlp_memory}
\end{figure}

We also distribute the layer normalization layer across \tprank{s} in memory-optimized distribution. To do this, observe that layer normalization is an element-wise operation given the mean and variance. Hence, we can decompose layer normalization into two steps, the first of which computes the local first and second moments of the data ($\sum_{i=1}^N x_i$, $\sum_{i=1}^N x_i^2$), which are then allreduced within the {\tpgroup} (note that this introduces minimal communication overhead since the moments are scalars) to compute the global mean and variance. Given the mean and variance, \cd{ApplyLayerNorm} operation applies element-wise normalization on the sharded activations. See Figures~\ref{fig:attention_memory} and \ref{fig:mlp_memory} for details.

Note that this design minimizes the redundancy in activation storage, since the input to every channel is fully sharded (and not replicated) across channels. This comes at the expense of increased communication, which results in four reduce-scatter operations and two scalar allreduce operations in the forward pass, and four allgather operations in the backward pass, per transformer layer.

\section{Communication Backend Architecture}
\label{sec:ap_backend}
\subsection{Overview}
When a Python object is to be communicated across {\pprank}s, the library first traverses the object to extract the tensors contained in it, and replaces them with a \cd{TensorStub} that contains an ID that is associated with the tensor. The object that is stripped of tensors is then serialized and communicated via MPI, between the CPU processes. The tensors are communicated through the backend as will be described in this section. After both the tensors and the encapsulating object are received at the destination, the receiver rank traverses the object, and replaces the \cd{TensorStub}s with the corresponding tensors based on the IDs they contain, to reconstruct the original object.

The communication backend consists of a set of threads, each having a specific responsibility, exchanging \cd{Message} objects between each other. A \cd{Message} contains all the necessary information regarding the communication of the tensor to be sent or received, such as the data pointers, shape, \cd{dtype}, device, peer rank (source or destination) for the communication, members for tracking MPI and CUDA events. The backend operates by having each thread monitor an input queue of \cd{Message}s, process the next \cd{Message}, and then enqueue it for another thread, depending on the next operation that needs to be performed on the \cd{Message}. Such design based on \cd{Message}-consumer threads allows flexibility in design and reduces complexity and coupling across components.

\begin{figure}[t]
\centering
\includegraphics[scale=0.5]{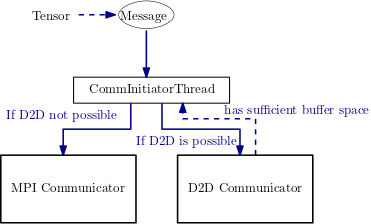}
\caption{System diagram for initial routing of messages for tensor communication}
\label{fig:d2d_top}
\end{figure}

When a tensor is to be sent or received, its metadata is encapsulated in a new \cd{Message} object, and enqueued at \cd{CommInitiatorThread}, which is responsible for routing incoming messages to one of two subsytems: MPI communicator and D2D communicator. When a \cd{Message} first arrives, this thread first initiates a round of metadata communication between source and destination, which contains information about shape, \cd{dtype}, and device of the tensor (this metadata communication is skipped for static mode, since it is the same in every step). This allows the receiver to allocate a correctly-sized buffer for the incoming tensor. The metadata communication takes place over the MPI communicator. When the metadata communication is complete, the corresponding \cd{Message} is routed back to \cd{CommInitiatorThread} for data communication.

If any of the following is true, then the message is routed to MPI communicator; otherwise D2D communicator is chosen:
\begin{itemize}
\item The incoming tensor device is CPU,
\item The source and destination ranks are on the same node, but there is no NVLink connecting them,
\item The source and destination ranks are on different nodes, but RDMA is not supported for the instance,
\item There is not enough free space in the D2D send or receive buffers (see Section~\ref{sec:d2d}) for the incoming tensor.
\end{itemize}
For performance, D2D communicator is prioritized whenever possible, since it avoids host-device copies over the PCIe link, and uses RDMA and NVLink technologies, for intra-node and cross-node communications, respectively. MPI communicator is used as a fallback whenever D2D is not feasible, as in the conditions listed above. We will next describe the architectures of MPI and D2D communicators in more detail.

\begin{figure}[t]
\centering
\includegraphics[scale=0.35]{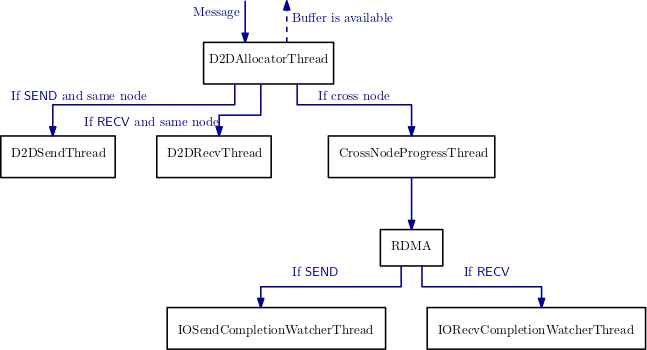}
\caption{System diagram for D2D Communicator block}
\label{fig:d2d}
\end{figure}

\subsection{Device-to-device communication} \label{sec:d2d}
D2D Communicator is responsible for direct copies between GPU devices, over NVLinks for devices on the same node, and over GPUDirect RDMA for devices in different nodes. 

A critical feature of this subsystem is that it needs to maintain persistent buffers for reception (and for cross-node D2D, for transmission) of tensors. This is because both cross-process CUDA memory copies, and RDMA transmissions, require creation of cross-process memory handles for the destination locations, which is an expensive operation. To amortize for this cost, we create persistent buffers for transmission at the beginning of program, which is re-used across different tensor transmissions (a memory manager is used to efficiently re-use this buffer). A side effect of this is that, an incoming tensor transmission must first check if there is sufficient space in the transmission buffers, and avoid using D2D communicator if there is not.

It consists of the following threads:
\begin{itemize}
\item {\bf \cd{D2DAllocatorThread}: } Responsible for MPI-based control signaling between source and destination, as well as allocation of sufficient space in destination buffers (and for cross-node D2D, also in source buffer). When this thread finishes processing a \cd{Message}, the source/destination transmission buffers are allocated, or if there is no sufficient space, the \cd{CommInitiatorThread} has been notified of this, so that the \cd{Message} can be routed over the MPI communicator. Crucially, the source and destination rank are in agreement about the outcome through a handshaking protocol.
\item {\bf \cd{D2DSendThread}: } For same-node D2D transmissions, initiates cross-GPU CUDA memory copies, given a destination memory handle (provided by \cd{D2DAllocatorThread}), and finalizes completed transmissions.
\item {\bf \cd{D2DRecvThread}: } For same-node D2D transmissions, tracks the reception of cross-GPU CUDA memory copies. For completed transmissions, copies the result to the framework tensor buffer, and notifies the framework.
\item {\bf \cd{CrossNodeD2DProgressThread}: } Responsible for making send/receive calls over the \cd{rdma-core} API (accessed through \emph{Herring} \cite{ThangakrishnanCavdar_20}), copying the incoming framework tensor to the transmission D2D buffer, and copying the received tensors to the newly-created framework tensor buffer.
\item {\bf \cd{IOSendCompletionWatcherThread}:} Tracks the completion of RDMA transmissions.
\item {\bf \cd{IORecvCompletionWatcherThread}:} Tracks the completion of RDMA receptions.
\end{itemize}

\begin{figure}[t]
\centering
\includegraphics[scale=0.4]{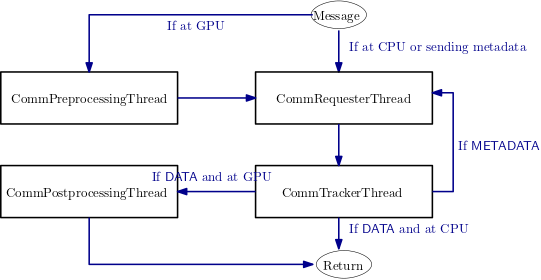}
\caption{System diagram for MPI Communicator block}
\label{fig:mpi_comm}
\end{figure}

\subsection{MPI-based communication}
MPI Communicator uses asynchronous, point-to-point MPI primitives to communicate tensors across CPU locations (within or across nodes). The system diagram for this component is given in Figure~\ref{fig:mpi_comm}, which consists of four main threads. We describe the functionality of each below:
\begin{itemize}
\item {\bf \cd{CommRequesterThread}:} Initiates the communication by calling the necessary asynchronous MPI primitives (\cd{MPI\_Isend}, \cd{MPI\_Irecv}). Always assumes that the source buffer is at CPU RAM. Attaches the resulting request (\cd{MPI\_Request} object) to the \cd{Message}, and enqueues it with \cd{CommTrackerThread}, which keeps track of ongoing communication transactions.
\item {\bf \cd{CommPreprocessingThread}:} If the original buffer is located in GPU, copies it to the CPU, updates the relevant fields of \cd{Message}, and enqueues it with \cd{CommRequesterThread}.
\item {\bf \cd{CommTrackerThread}:} Keeps track of ongoing tensor communications, finalizes those that are finished, and notifies the framework.
\item {\bf \cd{CommPostprocessingThread}:} For data transmissions (not metadata), if the tensor device is GPU, copies the received buffer to GPU, finalizes communication, and notifies the framework.
\end{itemize}


\section{Other Memory-Saving Features}
\label{sec:ap_other}
The library includes a number of crucial memory-saving techniques in support of the core model parallelism features, which are often needed for large-scale training. These techniques can be enabled through simple APIs, without requiring additional code changes.

\subsection{Optimizer state sharding}
Optimizer state sharding is a technique that was introduced by DeepSpeed, in \cite{RajbhandariRasley_20} (which was later extended to gradient and parameter sharding in \cite{RajbhandariRuwase_21}). It modifies the data parallelism workflow so that the optimizer states are sharded across data-parallel ranks. Instead of allreduce, the gradients are only reduced at the rank that stores the optimizer state for the corresponding parameter. This rank then locally updates these parameters, and broadcasts the latest values of the parameters to the rest of the ranks. This technique is readily implemented in the library, and can be enabled by setting \cd{"shard\_optimizer\_state": True} in model parallelism configuration (see Appendix~\ref{sec:api_config} for details).

\subsection{Activation checkpointing}
Activation checkpointing is another useful technique that drastically reduces the activation memory footprint during training. This is achieved by storing only a subset of activations (called ``checkpoints") during forward pass, and then re-computing the necessary subset of activations from the nearest checkpoint during backward pass. The library offers a fine-grained API for activation checkpointing, where the user can choose exactly which modules to checkpoint. For \cd{nn.Sequential} modules, the user can also specify the frequency with which checkpointing should occur, in number of layers. This feature can be enabled on PyTorch by using \cd{smp.set\_activation\_checkpointing} API (see Appendix~\ref{sec:api_ckpt} and \ref{sec:api_config} for API details)

\subsection{Activation offloading}
Activation offloading is yet another useful memory-saving feature, where the stored activations are offloaded to CPU RAM during forward pass, and fetched back to the GPU during backward pass, when they are needed. In SageMaker model parallelism, this feature is implemented in a way that interacts with, and requires the use of, pipeline parallelism and activation checkpointing. Specifically, the activations for a particular microbatch are fetched before the backward execution for that microbatch, and the activations for the other microbatches remain in CPU RAM. Moreover, it is only the checkpointed activations that are offloaded, \emph{i.e.}, short-term recomputed activations are not offloaded. In order to prevent the backward computation getting blocked on the fetching of activations from CPU over the PCIe link, in practice we prefetch the activations, by starting the data transfer shortly \emph{before} the backward computation starts. The amount of head-start can be controlled through the configuration parameter \cd{"activation\_loading\_horizon"}, and activation offloading itself can be enabled through \cd{"offload\_activations"} (see Appendix~\ref{sec:api_config} for details).

\section{Detailed API documentation}
\label{sec:ap_api}
\subsection{Additional configuration parameters}\label{sec:api_config}
The following configuration parameters are available in addition to the documentation provided in \cite{smp_apidoc}. 
\begin{itemize}
\item \cd{tensor\_parallel\_degree} (\cd{int}, default: 1) The number of devices over which the tensor-parallel modules will be distributed. If \cd{tensor\_parallel\_degree} is more than 1, then \cd{ddp} must be set to \cd{True}.
\item \cd{pipeline\_parallel\_degree} (\cd{int}, default: 1) The number of devices over which pipeline parallelism will be performed (alias for \cd{partitions})
\item \cd{optimize} (\cd{["speed", "memory"}], default: \cd{"memory"}) Determines the distribution mechanism of transformer layers. If optimizing speed, there will be less communication across tensor-parallel ranks and layer normalization will not be distributed, but there will be duplicate activations stored across tensor-parallel ranks. If optimizing memory, there will be no redundant activations stored, but this will result in more communication overhead across tensor-parallel ranks.
\item \cd{fp16\_params} (\cd{bool}, default: \cd{False}) If \cd{True}, the parameters of the distributed modules will be initialized in \cd{fp16}.
\item \cd{shard\_optimizer\_state} (\cd{bool}, default: \cd{False}) If \cd{True}, shards the optimizer state of all parameters across the data parallel processes which hold the same parameter. This sharding of optimizer state happens in a balanced manner. 
\item \cd{offload\_activations} (\cd{bool}, default: \cd{False}) If \cd{True}, offloads the checkpointed activations to CPU, and fetches them back before the backward pass of the corresponding microbatch.
\item \cd{activation\_loading\_horizon} (\cd{int}, default: 4) If activation offloading is enabled, determines how early the activations should be brought back to the GPU. If too small, might impact performance by blocking the GPU on the CPU-GPU data transfers. If too large, might increase GPU memory usage.
\item \cd{\_prescaled\_batch} (\cd{bool}, default: \cd{False}) If \cd{True}, when \cd{DistributedTransformerLMHead} is used (this is typically used for GPT-2/3), the library assumes that the devices in the same tensor parallelism group receive the same input data. Otherwise, it is assumed that they receive different examples 
\item \cd{placement\_strategy} (\cd{str}, default: \cd{"cluster"}) Determines the mapping of model partitions onto physical devices.
\begin{itemize}
    \item Must be either \cd{"spread"}, \cd{"cluster"}, or a permutation of the string \cd{"DPT"}. \cd{"spread"} is equivalent to \cd{"TPD"}, and \cd{"cluster"} is equivalent to \cd{"DPT"}. The interpretation of the three-letter string is as follows:
    \item \cd{D} stands for (reduced) data parallelism, \cd{P} stands for pipeline parallelism, and \cd{T} stands for tensor parallelism.
    \item As one moves right-to-left on the three-letter string, the parallelism type that is represented by the letter type is performed over global ranks that are closer together. The parallelism type that is represented by the right-most letter is performed over immediate-neighbor ranks (i.e., devices), while the parallelism type represented by the left-most letter is performed over ranks that are as distant as possible.
    \end{itemize}
\end{itemize}

\subsection{Enabling activation checkpointing}\label{sec:api_ckpt}
\cd{smp.set\_activation\_checkpointing}
\begin{itemize}
    \item This API enables checkpointing for a module given a reference to the module. 
    \item {\bf Arguments}:
    \begin{itemize}
        \item \cd{module} (Instance of \cd{nn.Module} or \cd{nn.Sequential}): The module to checkpoint.
        \item \cd{preserve\_rng\_state} (\cd{bool}, default=\cd{True}): Set to False to omit stashing and restoring the RNG state during each checkpoint. 
        \item \cd{pack\_args\_as\_tuple} (\cd{bool}, default=\cd{False}): Can only be passed when module is a sequential module. To ensure that backward works correctly, the autograd function has to unpack any tuples received. If the layer checkpointed takes a tuple as input, then this needs to be set to \cd{True}.
        \item \cd{strategy}: (string, default=\cd{“each”}): Can only be passed when module is a sequential module. Strategy determines how many layers part of the sequential module need to be grouped together for one checkpointing call. 
        \item  This determines how much memory can be reduced. It can take the following values
        \begin{itemize}
            \item \cd{"each"} : The default is to checkpoint each module inside the sequential separately. 
            \item \cd{"contiguous"}: Groups consecutive layers on the same partition together. For example if a sequential consists of \cd{[a, b, c, d]} where \cd{a, b} are on \pprank{ }0 and \cd{c, d} are on \pprank{ }1, then this strategy would checkpoint \cd{a, b} together and then \cd{c, d} together. This means effectively, the inputs of \cd{a}, outputs of \cd{b}, inputs of \cd{c}, and outputs of \cd{d} are in memory, rest of the activations are recomputed. 
            \item \cd{"group\_2"}, \cd{"group\_3"}, \cd{"group\_4"}, etc: More generally, \cd{group\_x} where \cd{x} is an integer. This strategy provides more flexibility in how many layers to group together. \cd{group\_x} groups \cd{x} layers together on a best effort basis. It can group \cd{x} layers together if there are \cd{x} layers consecutively on the same partition. For example: \cd{[a,b,c,d,e]} where \cd{a, b} are on \pprank{ }0 and \cd{c, d, e} are on \pprank{ }1. If the strategy is \cd{group\_3}, then \cd{a, b} are checkpointed together on \pprank{ }0 and \cd{c, d, e} are checkpointed together on \pprank{ }1.
            \end{itemize}
\end{itemize}
\end{itemize}

\subsection{\cd{smp.nn} module}\label{sec:api_nn}
The following are the built-in distributed modules that are part of \cd{smp.nn} module.
\begin{itemize}
\item \cd{smp.nn.DistributedLinear (in\_features, out\_features)}
\begin{itemize}
\item Distributed implementation for \cd{nn.Linear}
\end{itemize}
\item \cd{smp.nn.DistributedEmbedding (num\_embeddings, embedding\_dim, padding\_idx=None, max\_norm=None, norm\_type=2.0, scale\_grad\_by\_freq=False, sparse=False, \_weight=None, initializer\_range=0.02, \_skip\_allgather=False, \_skip\_scatter\_and\_merge=False,)}
\begin{itemize}
\item Distributed implementation for \cd{nn.Embedding}
\end{itemize}
\item \cd{smp.nn.DistributedTransformerLMHead (num\_layers=12, num\_attention\_heads=32, attention\_head\_size=32, hidden\_size=1024, intermediate\_size=4096, vocab\_size=30522, num\_positions=1024, attention\_dropout\_prob=0.1, hidden\_dropout\_prob=0.1, activation="gelu", layernorm\_epsilon=1e-5, num\_token\_types=0, causal\_mask\_size=None, add\_cross\_attention=False, add\_lm\_head=True,  initializer\_range=0.02, use\_normal\_initialization=False, pre\_layernorm=False, post\_layernorm=True)}
\begin{itemize}
\item Distributed implementation for GPT-3, including the embedding and LM head
\end{itemize}
\item \cd{smp.nn.DistributedTransformer (num\_layers=12, num\_attention\_heads=32, attention\_head\_size=32, hidden\_size=1024, intermediate\_size=4096, attention\_dropout\_prob=0.1, hidden\_dropout\_prob=0.1, activation="gelu", layernorm\_epsilon=1e-5, initializer\_range=0.02, use\_normal\_initialization=False, causal\_mask\_size=None, add\_cross\_attention=False, pre\_layernorm=False, post\_layernorm=True)}
\begin{itemize}
\item Distributed implementation for a sequence of generic transformer layers, which can be specialized to BERT, GPT-2/3, RoBERTa, and many other transformers.
\end{itemize}
\item \cd{smp.nn.DistributedTransformerLayer (num\_attention\_heads=32, attention\_head\_size=32, hidden\_size=1024, intermediate\_size=4096, attention\_dropout\_prob=0.1, hidden\_dropout\_prob=0.1, activation="gelu", layernorm\_epsilon=1e-5, initializer\_range=0.02, use\_normal\_initialization=False, causal\_mask\_size=None, add\_cross\_attention=False, pre\_layernorm=False, post\_layernorm=True)}
\begin{itemize}
\item Distributed implementation for a single generic transformer layer.
\end{itemize}
\end{itemize}

\subsection{Registering distributed modules}\label{sec:api_register}
\begin{itemize}
\item \cd{@smp.tp\_register (dist\_module, init\_hook=None, forward\_hook=None, return\_hook=None)}
\begin{itemize}
    \item A class decorator that registers the \cd{dist\_module} class with the module class that it is attached to. The hooks can be used to adapt to different interfaces used with \cd{\_\_init\_\_} and \cd{forward} methods.
    \item {\bf Arguments: }
    \begin{itemize}
        \item \cd{dist\_module}: A subclass of \cd{smp.nn.DistributedModule} that implements the distributed version of the module class the decorator is attached to. Any distributed module class defined in \cd{smp.nn} module can be used.
        \item \cd{init\_hook}: A callable that translates the arguments of the original module \cd{\_\_init\_\_} method to an (\cd{args}, \cd{kwargs}) tuple compatible with the arguments of the corresponding distributed module \cd{\_\_init\_\_} method. Must return a tuple, whose first element is an iterable representing the positional arguments, and second element is a dict representing the keyword arguments. The input signature of the \cd{init\_hook} must \emph{exactly} match the signature of the original \cd{\_\_init\_\_} method (including argument order and default values), except it must exclude self.
        \item \cd{forward\_hook}: A callable that translates the arguments of the original module forward method to an (\cd{args}, \cd{kwargs}) tuple compatible with the arguments of the corresponding distributed module forward method. Must return a tuple, whose first element is an iterable representing the positional arguments, and second element is a dict representing the keyword arguments. The input signature of the \cd{init\_hook} must {\bf exactly} match the signature of the original forward method (including argument order and default values), except it must exclude self.
        \item \cd{return\_hook}: A callable that translates the object returned from the distributed module to the return object expected of the original module.
\end{itemize}
\end{itemize}
\item \cd{smp.tp\_register\_with\_module (module\_cls, dist\_module, init\_hook=None, forward\_hook=None, return\_hook=None)}
\begin{itemize}
    \item When there is no direct access to model definition code, this API can be used to register a distributed module with an existing module class. 
    \item {\bf Arguments:}
    \begin{itemize}
        \item \cd{module}: The existing module class to be distributed
        \item \cd{dist\_module}: A subclass of \cd{smp.nn.DistributedModule} that implements the distributed version of the module class the decorator is attached to. Any distributed module class defined in \cd{smp.nn} module can be used.
        \item \cd{init\_hook}: A callable that translates the arguments of the original module \cd{\_\_init\_\_} method to an (\cd{args}, \cd{kwargs}) tuple compatible with the arguments of the corresponding distributed module \cd{\_\_init\_\_} method. Must return a tuple, whose first element is an iterable representing the positional arguments, and second element is a dict representing the keyword arguments. The input signature of the \cd{init\_hook} must \emph{exactly} match the signature of the original \cd{\_\_init\_\_} method (including argument order and default values), except it must exclude self.
        \item \cd{forward\_hook}: A callable that translates the arguments of the original module forward method to an (\cd{args}, \cd{kwargs}) tuple compatible with the arguments of the corresponding distributed module forward method. Must return a tuple, whose first element is an iterable representing the positional arguments, and second element is a dict representing the keyword arguments. The input signature of the \cd{init\_hook} must {\bf exactly} match the signature of the original forward method (including argument order and default values), except it must exclude self.
        \item \cd{return\_hook}: A callable that translates the object returned from the distributed module to the return object expected of the original module.
\end{itemize}
\end{itemize}
\end{itemize}

\subsection{Delaying parameter initialization}\label{sec:api_delay}
\begin{itemize}
\item \cd{smp.delay\_param\_initialization:} A context manager that delays the CPU initialization of \cd{nn.Parameter}s until the modules are moved to GPU. Any modules created inside the context manager will not have their parameters taking up physical memory, until the first call of a \cd{smp.step}-decorated function. Useful for cases when the model parameters are too many to fit in the CPU RAM.
\end{itemize}

\subsection{Creating custom distributed modules}\label{sec:api_custom}
The following API can be used to create custom modules, after importing them from \cd{smdistributed.modelparallel.torch.nn.utils}. All custom distributed modules must be sub-class of \cd{smp.nn.DistributedModule}.
\begin{itemize}
\item \cd{parameter\_creation\_scope (module, scaled\_batch=True, dtype=None, use\_normal=False, initializer\_range=0.02)}
\begin{itemize}
\item Parameters of the \cd{smp.nn.DistributedModule} must be created within this context manager. \cd{module} is a reference to the module object \cd{self}, \cd{scaled\_batch} signifies whether the parameter interacts with the entire batch collected from \tpgroup, \cd{use\_normal} initializes parameters with normal distribution with range \cd{initializer\_range}.
\end{itemize}
\item \cd{initialize\_with\_input\_partition (module)}
\begin{itemize}
\item A context manager to create parameters that are distributed across their input channels, \emph{i.e.}, the dimension that is applied to the input tensor. \cd{module} is a reference to the module \cd{self} object.
\end{itemize}
\item \cd{initialize\_with\_output\_partition}
\begin{itemize}
\item A context manager to create parameters that are distributed across their output channels. \cd{module} is a reference to the module \cd{self} object.
\end{itemize}
\item \cd{fused\_allgather\_for\_tp (tensor, dim)}
\begin{itemize}
\item Applies allgather collective to tensors across \tpgroup, and concatenates the result across dimension \cd{dim}.
\end{itemize}
\item \cd{fwd\_allreduce\_for\_tp (tensor)}
\begin{itemize}
\item Applies allreduce collective to tensors across \tpgroup.
\end{itemize}
\item \cd{scatter\_and\_merge\_for\_tp (tensor, split\_dim, merge\_dim)}
\begin{itemize}
\item Slices tensors across \cd{split\_dim} into as many slices as the tensor parallelism degree, applies all-to-all to resulting slices, and concatenates the received slices across \cd{merge\_dim}.
\end{itemize}
\item \cd{bwd\_allreduce\_for\_tp (tensor)}
\begin{itemize}
\item Applies allreduce collective to tensors across \tpgroup during backward pass (no-op for forward pass).
\end{itemize}
\item \cd{reduce\_scatter\_for\_tp (tensor, dim)}
\begin{itemize}
\item Slices \cd{tensor} across \cd{dim}, applies reduce-scatter collective to the slices across \tpgroup.
\end{itemize}
\end{itemize}


\end{document}